\documentclass{article}

\PassOptionsToPackage{numbers,sort&compress}{natbib}
\usepackage[main,final]{neurips_2026}

\usepackage[utf8]{inputenc}
\usepackage[T1]{fontenc}
\usepackage{hyperref}
\usepackage{url}
\usepackage{booktabs}
\usepackage{amsfonts}
\usepackage{nicefrac}
\usepackage{microtype}
\usepackage[svgnames,table]{xcolor}
\usepackage{graphicx}
\usepackage{amsmath}
\usepackage{amssymb}
\usepackage{mathtools}
\usepackage{multirow}
\usepackage{adjustbox}
\usepackage{makecell}
\usepackage{colortbl}
\usepackage{tabularx}
\usepackage{ragged2e}
\usepackage{tcolorbox}
\usepackage{tikz}
\usepackage{afterpage}
\hypersetup{hypertexnames=false,hidelinks}

\definecolor{tableRule}{HTML}{6D8791}
\definecolor{tableHeader}{HTML}{EAF3F5}
\definecolor{tableHighlight}{HTML}{EAF7EE}
\definecolor{tableSoft}{HTML}{FFF6E6}
\definecolor{tableWarn}{HTML}{FFF0E8}
\AtBeginDocument{\arrayrulecolor{tableRule}}
\newcommand{\maintablefont}{\footnotesize\setlength{\tabcolsep}{2.6pt}\renewcommand{\arraystretch}{0.94}}
\newcommand{\appendixtablefont}{\footnotesize\setlength{\tabcolsep}{3.2pt}\renewcommand{\arraystretch}{0.98}}
\newcommand{\appendixwidetablefont}{\scriptsize\setlength{\tabcolsep}{2.8pt}\renewcommand{\arraystretch}{0.96}}
\newcommand{\tablehead}{}
\newcommand{\methodrow}{\rowcolor{tableHighlight}}
\newcommand{\alternativerow}{\rowcolor{tableSoft}}

\newcommand{\E}{\mathbb{E}}
\newcommand{\KLhat}{\widehat{\mathcal{D}}_{\mathrm{KL}}}
\newcommand{\clip}{\mathrm{clip}}
\newcommand{\Hent}{\mathcal{H}}

\newcolumntype{Y}{>{\RaggedRight\arraybackslash}X}
\newcolumntype{L}[1]{>{\RaggedRight\arraybackslash}p{#1}}

\title{IRPO: Boosting Image Restoration via Post-training GRPO}

\author{
\normalfont Haoxuan Xu$^{1,2,\dagger}$\quad
Yi Liu$^{4,\dagger}$\quad
Tianfu Li$^{1}$\quad Ruolin Shen$^{5}$\quad Boyuan Jiang$^{6}$\\
Jinlong Peng$^{7}$\quad Donghao Luo$^{8}$\quad
Xiaobin Hu$^{3}$\thanks{Corresponding author.}\quad
Shuicheng Yan$^{3}$\quad Haoang Li$^{1}$\\[3pt]
\small $^{1}$The Hong Kong University of Science and Technology (Guangzhou)\\
\small $^{2}$Shenzhen Loop Area Institute\quad $^{3}$National University of Singapore\\
\small $^{4}$Tsinghua University\quad $^{5}$Technical University of Munich\\
\small $^{6}$Zhejiang University\quad $^{7}$Shanghai Jiao Tong University\quad $^{8}$Fudan University
}
\begin{document}
\raggedbottom

\maketitle
{\renewcommand{\thefootnote}{\fnsymbol{footnote}}
\footnotetext[2]{Haoxuan Xu and Yi Liu contributed equally.}}

\begin{abstract}
Post-training has become effective for high-level generation, but its role in low-level vision remains underexplored. Existing image restoration methods often rely on fixed pixel-wise fitting to ground-truth images, which can lead to over-smoothing and weak generalization. We propose IRPO, a GRPO-based post-training framework for deterministic restoration models. IRPO is built around two axes: data formulation and reward modeling. For data formulation, we select the 30\% underperforming samples from the pre-training stage, which improves both accuracy and training efficiency. For reward modeling, we combine fidelity-oriented and quality-aware feedback with three components: a General Reward for structural fidelity, an Expert Reward that uses a Vision-Language Model as a coarse visual-quality judge, and a Restoration Reward for task-specific low-level cues. Experiments on six in-domain and five out-of-domain (OOD) benchmarks show that IRPO improves the AdaIR baseline by 0.93 dB on in-domain tasks and 3.43 dB on OOD settings.
\end{abstract}

\afterpage{%
\begin{figure}[t]
    \centering
    % \begin{tikzpicture}
    {\includegraphics[width=8.6cm]{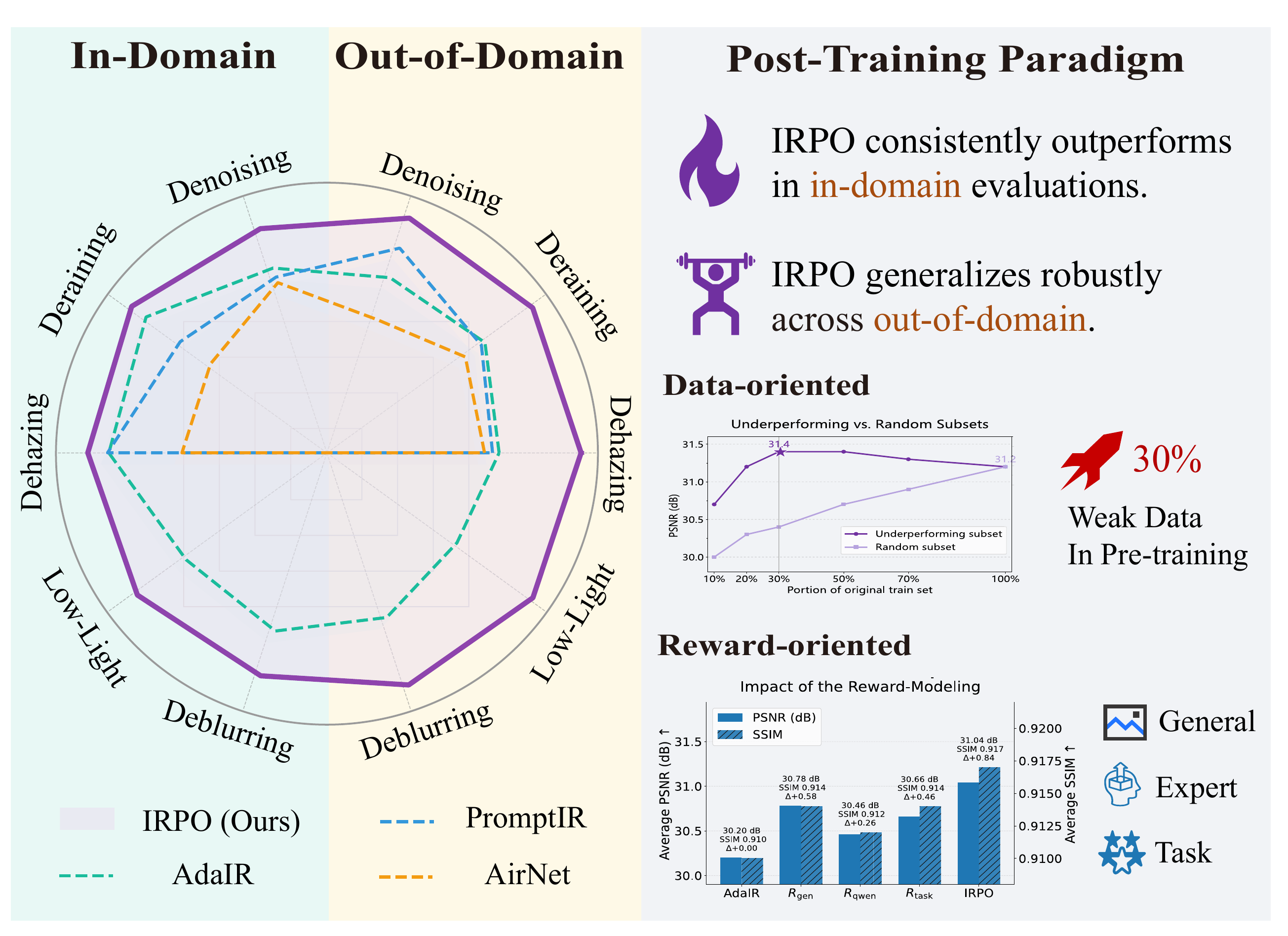}}
    % \fill[black!4] ([xshift=5.18cm,yshift=5.22cm]img.south west) rectangle ([xshift=5.69cm,yshift=5.54cm]img.south west);
    % \node[anchor=west, font=\fontsize{6.2}{6.2}\selectfont\bfseries] at ([xshift=5.20cm,yshift=5.37cm]img.south west) {IRPO};
    % \end{tikzpicture}
    \caption{Overview of IRPO and its empirical behavior. 
    (Left) Average PSNR comparison across in-domain and OOD settings. (Right) The two core components of IRPO: data-oriented supervision on the 30\% underperforming subset, and reward-oriented optimization with complementary restoration rewards.}
    \label{fig:intro_summary_figure}
    \vspace{-4pt}
\end{figure}%
}

\section{Introduction}
\label{sec:intro}
%图：（1）首图：引入咱解决什么问题（data）。
% （2）框图：再优化优化
%（3）折现图：反映data特征的
% （4）loss图：拼在一起的
% （5）可视化的图：放大一下 看出差异。
Recent generative models cover diverse tasks, including virtual try-on with VTON-HandFit~\citep{liang2025vton}, portrait animation with Sonic~\citep{ji2025sonic}, and controllable 3D asset generation with Hunyuan3D-Omni~\citep{post-training-3d}. Post-training offers a way to refine pre-trained models using selected data and reward signals~\citep{post-training-define}. For example, DDPO~\citep{black2023ddpo} optimizes image generation against downstream rewards, while T2V-Turbo-v2~\citep{post-training-video2} combines data selection, reward feedback, and conditional guidance for video generation. Adversarial post-training also improves efficient video generation~\citep{post-training-video,post-training-video3}.
% This process enhances perceptual quality and semantic consistency, bridging the gap between statistical fidelity and subjective realism~\cite{post-training-Preference}. 
The progress of post-training in high-level tasks~\citep{black2023ddpo,post-training-video2} motivates a related question for low-level vision. Image restoration (IR) aims to recover faithful, natural-looking images from degraded inputs. Quality-aware approaches include fine-grained image quality assessment for perceptual restoration~\citep{low-Perceptual} and IQA-driven RL for mixed-distortion restoration~\citep{low-Perceptual2}. The key question is how to adapt post-training to IR, where paired supervision remains essential but the optimization signal should better handle hard cases and real-world degradations.

% ("current low level training paradigm + problem $==>$ a novel training paradigm in low level")
% Image restoration (IR) is a fundamental task in computer vision, which has been widely used in numerous downstream applications across diverse domains, including surveillance, medical imaging, and remote sensing. 
Existing restoration methods employ different architectural and training choices. \textbf{1) End-to-end supervised learning}: a network learns a mapping from degraded inputs to clean targets~\citep{single}. \textbf{2) Multi-stage designs}: MPRNet and Diff-Mamba use successive restoration stages~\citep{MPRNet,Diff-Mamba} to refine restored outputs. \textbf{3) Domain-specific modeling and adaptation}: methods such as APISR for anime super-resolution~\citep{apisr} and MoA-VR for video restoration~\citep{video_fine} tailor models to their target domains. These choices provide complementary ways to improve restoration; our focus is how to post-train an existing restoration model using selected paired data and reward feedback.
Recent work also explores training-time privileged information and iterative refinement for all-in-one restoration, as in SIPL~\citep{wu2025boosting}. More broadly, minimizing distortion alone does not ensure high perceptual quality, as formalized by the perception--distortion tradeoff~\citep{quality}. This motivates studying complementary optimization signals for restoration.
% Overall, these low-level training paradigms follow a deterministic supervision framework~\cite{wu2025boosting} that hard-fits predefined static objectives, lacking dynamic behavioral regulation mechanisms. Such hard-fitting supervision training hinders the formation of a post-hoc adaptive policy alignment loop, failing to effectively align the model’s behavior with human perceptual preferences~\cite{quality} and thereby limiting the model’s generalization and adaptability under complex low-level degradations and real-world conditions.

% Overall, these training strategies, which rely solely on L1 or L2 losses for hard-fitting to ground-truth images at the pixel level~\cite{perceptual}, can lead to over-smoothing and distortion~\cite{quality}. These objectives fail to capture perceptual preference, semantic consistency, and user-intent alignment, thereby limiting the overall restoration quality. 
% Therefore, inspired by recent post-training advances in image generation~\cite{post-training-image}, a pressing question arises: Can we design a universal post-training paradigm for low-level task that is model-agnostic, requires no additional training data, inherently mitigates overfitting, and enables efficient plug-and-play fine-tuning of restoration models?

% \textit{(why "GRPO post-training for low level" + the chanllenge in "GRPO post-training for low level": data reward)}
These limitations motivate a post-training objective beyond fixed pixel losses. Many useful restoration criteria, such as perceptual similarity, visual plausibility, and degradation-specific quality, cannot be reduced to a single differentiable pixel loss. Reinforcement Learning (RL), including Group Relative Policy Optimization (GRPO)~\citep{grpo}, offers a framework for reward-based policy optimization. We investigate its application to IR post-training. The design needs two components. \textbf{1) Data formulation}: unlike high-level tasks where multiple outputs can be ranked by preference, IR requires a stable rule for selecting samples that are worth post-training across degradations and domains. Paired data are also expensive to collect~\citep{data}, while synthetic degradations may not cover real-world cases. \textbf{2) Reward modeling}: perceptual losses provide feature-based supervision beyond pixel-wise error~\citep{perceptual}. Our goal is to complement supervision with rewards for structural fidelity, visual quality, and degradation-specific residuals.

% \textit{(Our Methods to address the chanllenge)}
To address these issues, we propose \textbf{I}mage \textbf{R}estoration with GR\textbf{PO}-based post-training (IRPO), a framework built on data-oriented supervision and reward-oriented optimization (Fig.~\ref{fig:intro_summary_figure}).
We first study data formulation through random sampling and performance-based selection. This leads to a simple data-oriented principle: post-training on the 30\% underperforming samples from the pre-training stage gives a favorable balance between performance and efficiency.
% we explore data formulation and discover a simple yet effective principle:
 % We study data formulation via extensive random sampling and performance-based selection, leading to
 % a simple yet effective rule: post-training exclusively on the 30\% underperforming samples from the pre-training stage yields the optimal balance of performance and efficiency.
% This data-oriented post-training keeps a balanced amount of hard samples to improve gradient use and distribution coverage, boosting the IR model’s representation without extra data.
% To be specific, we first explore the data formulation for low-level post-training paradigm through extensive random sampling and performance-based selection across different data configurations. Building upon these explorations, we then formulate a simple yet effective data-oriented principle for the low-level post-training paradigm, where using the underperforming samples from the pre-training stage as post-training data consistently yields more generalizable and accurate performance. 
% This data-oriented supervision of the low-level post-training paradigm maintains a moderate proportion of challenging samples to maximize gradient efficiency and distributional coverage, thereby enhancing the representational capacity of the IR model without additional training data.
Next, low-level tasks differ substantially in degradation type, and their optimization objectives must preserve fidelity while also capturing quality cues that are not fully reflected by a single pixel loss.
To this end, we build a reward system with three parts: (1) a General Reward for structural fidelity and broad quality; (2) an Expert Reward that uses a Vision-Language Model (VLM) as a coarse visual-quality judge; and (3) a Restoration Reward for degradation-dependent low-level cues. Finally, motivated by all-in-one IR~\citep{cui2025adair, promptir}, we evaluate GRPO post-training under mixed-degradation training to test its adaptability to real-world settings.

In summary, our contributions are as follows:

\begin{itemize}
\item We introduce a low-level GRPO-based post-training framework that jointly studies data formulation and reward modeling for image restoration.
\item We identify a performance-based data principle that directs post-training to hard samples without requiring additional data.
\item We design a reward-oriented optimization scheme that combines fidelity, coarse visual-quality feedback, and degradation-specific restoration cues.
\item We evaluate IRPO on six in-domain and five out-of-domain benchmarks, where it improves the AdaIR backbone and shows consistent robustness in real-world settings.

\end{itemize}

\section{Related Work}
\label{sec:formatting}
\subsection{Low-Level Training Paradigm}
% The evolution of deep learning has led to a variety of training strategies for low-level vision models, which can be broadly categorized into three paradigms:\\
Related restoration approaches differ in their architectures and adaptation strategies:
% 1) Single-stage Training. This is the most straightforward and widely adopted paradigm, where a model is trained end-to-end in a single phase to directly map degraded inputs to clean outputs. Pioneering works based on convolutional neural networks, such as DnCNN~\cite{DnCNN} and SRCNN~\cite{SRCNN}, exemplify this approach. Its simplicity and efficiency are further demonstrated in later transformer-based models like SwinIR~\cite{SwinIR} and Restormer~\cite{restormer}, which rely on powerful architectures to learn the restoration mapping. \\
1) Single-stage Training. Models are trained end-to-end to map degraded inputs to clean outputs. Early CNN-based restorers such as DnCNN~\citep{DnCNN} and SRCNN~\citep{SRCNN}, and later transformer variants like SwinIR~\citep{SwinIR} and Restormer~\citep{restormer}, demonstrate that strong architectures can learn the restoration mapping efficiently.
2) Multi-stage Designs. Multi-stage architectures such as MPRNet~\citep{MPRNet} and Diff-Mamba~\citep{Diff-Mamba} refine restoration through successive stages. Related multi-scale architectures, such as MIRNet-v2~\citep{mirnet}, aggregate features across resolutions. These designs can improve restoration quality, but introduce additional architectural or training complexity.
3) Domain-specific Modeling and Adaptation. Pre-trained models can be adapted to specific domains, as in FeRA~\citep{yin2025fera} for diffusion adaptation. Domain-specific restoration also uses tailored data and priors: CBDNet~\citep{cbdnet} targets real camera noise, ESRGAN~\citep{ESRGAN} perceptual super-resolution, and APISR~\citep{apisr} anime super-resolution. Other examples include 3D facial priors for face super-resolution and restoration~\citep{hu2020face,hu2021face}, and video restoration systems such as VRT~\citep{vrt} and MoA-VR~\citep{video_fine}. These approaches motivate adaptation to the target degradation, although their training data and objectives may not transfer across domains.
Restoration objectives must account for the perception--distortion tradeoff~\citep{quality}; perceptual losses offer one way to move beyond pixel-wise error~\citep{perceptual}. We study a GRPO-based post-training framework for low-level tasks. Its two core components are data-oriented supervision and reward-oriented optimization, which are designed to improve robustness under diverse real-world degradations.

% Traditional image restoration methods typically utilize specialized models designed for specific degradation types, which limits their applicability in real-world scenarios where images often suffer from multiple types of degradation simultaneously. Consequently, there is an increasing need for all-in-one models that can address numerous degradations within a single framework. 

% Early unified models employed specialized encoder and decoder heads to address distinct restoration tasks. Subsequent research focused on eliminating the need for prior degradation knowledge by developing conditioning mechanisms, which evolved from basic degradation encoders to advanced frameworks enhanced with frequency priors or dimensionality reduction. More recently, prompt-learning-based approaches have been introduced to overcome the limitations of pretrained prior-based methods. 

\vspace{-0.15cm}
\subsection{Reinforcement Learning}

% Recent advancements in RL have substantially enhanced large language models’ (LLMs) reasoning capabilities~\cite{llm2}\cite{llm}, particularly through the post-training paradigm that explicitly elicit deliberate, stepwise reasoning before producing final answers. For example, OpenAI's o1 model~\cite{o1} exemplifies this trend, demonstrating that reinforcement learning can substantially improve structured reasoning and problem-solving abilities. Unlike conventional RL with Human Feedback (RLHF)~\cite{RLHF} frameworks that rely on complex value networks, GRPO~\cite{grpo} achieves stable and efficient learning by leveraging intra-group relative advantage estimation. Building on this paradigm, DeepSeek-R1~\cite{deepseek} introduced a rule-based reward mechanism combined with GRPO, establishing a post-training paradigm that encourages models to conduct comprehensive reasoning prior to output generation. Owing to this efficiency and generality, GRPO has rapidly expanded beyond natural language processing to multimodal generation~\cite{R1,r12,r13}, speech understanding~\cite{understanding}, and decision-making tasks~\cite{vla}\cite{vln}, achieving superior performance across high-level domains.

RL-based post-training improves LLM reasoning in OpenAI's o1~\citep{o1} and DeepSeek-R1~\citep{deepseek}. RLHF~\citep{RLHF} commonly uses an actor--critic approach, whereas GRPO, introduced in DeepSeekMath~\citep{grpo}, estimates relative advantages within each sampled group without a separate value model. Direct Preference Optimization (DPO)~\citep{llm} instead learns from preference pairs without an explicit RL optimization loop. GRPO-style methods extend to multimodal reasoning in R1-V~\citep{R1}, R1-OneVision~\citep{r12}, and R1-VL~\citep{r13}, multimodal table understanding~\citep{understanding}, and action or navigation tasks in VLA-R1~\citep{vla} and VLN-R1~\citep{vln}. In related multimodal settings, SemMSA~\citep{li2026semmsa} and VT-FSL~\citep{li2025vtfsl} use semantic guidance for sentiment analysis and few-shot recognition, while DVLA-RL~\citep{li2026dvla} uses RL-based gating for vision--language alignment.

% This post-training paradigm has been widely adopted in high-level generation tasks~\cite{post-training-image}~\cite{post-training-video2}, where it serves as an alignment phase to produce outputs that more faithfully reflect human intent. 
% Despite this post-training paradigm successes in high-level tasks, the effectiveness of GRPO-based post-training remains largely underexplored in the low-level setting, due to the scarcity of low-level training data~\cite{data} and the complexity of reward objectives~\cite{quality}. Similar to high-level tasks, low-level tasks also require a delicate balance between physical fidelity and perceptual naturalness to restore more realistic images~\cite{low-Perceptual}. These insights motivate us to investigate a novel low-level GRPO-based training paradigm that systematically explores both data formulation and reward modeling, laying the foundation for more generalizable and accurate image restoration.
Despite these advances, GRPO-style post-training remains underexplored for low-level restoration. This setting raises questions about paired data~\citep{data}, deterministic restoration backbones, and suitable rewards. In particular, restoration rewards must balance perceptual quality and fidelity, consistent with the perception--distortion tradeoff~\citep{quality}. Reward optimization should therefore complement, rather than replace, supervised constraints. We study this setting through the two central questions of IRPO: data formulation and reward modeling.

\section{Methodology}
\label{sec:method}

\begin{figure*}[!t]
    \centering
    \includegraphics[width=0.99\linewidth]{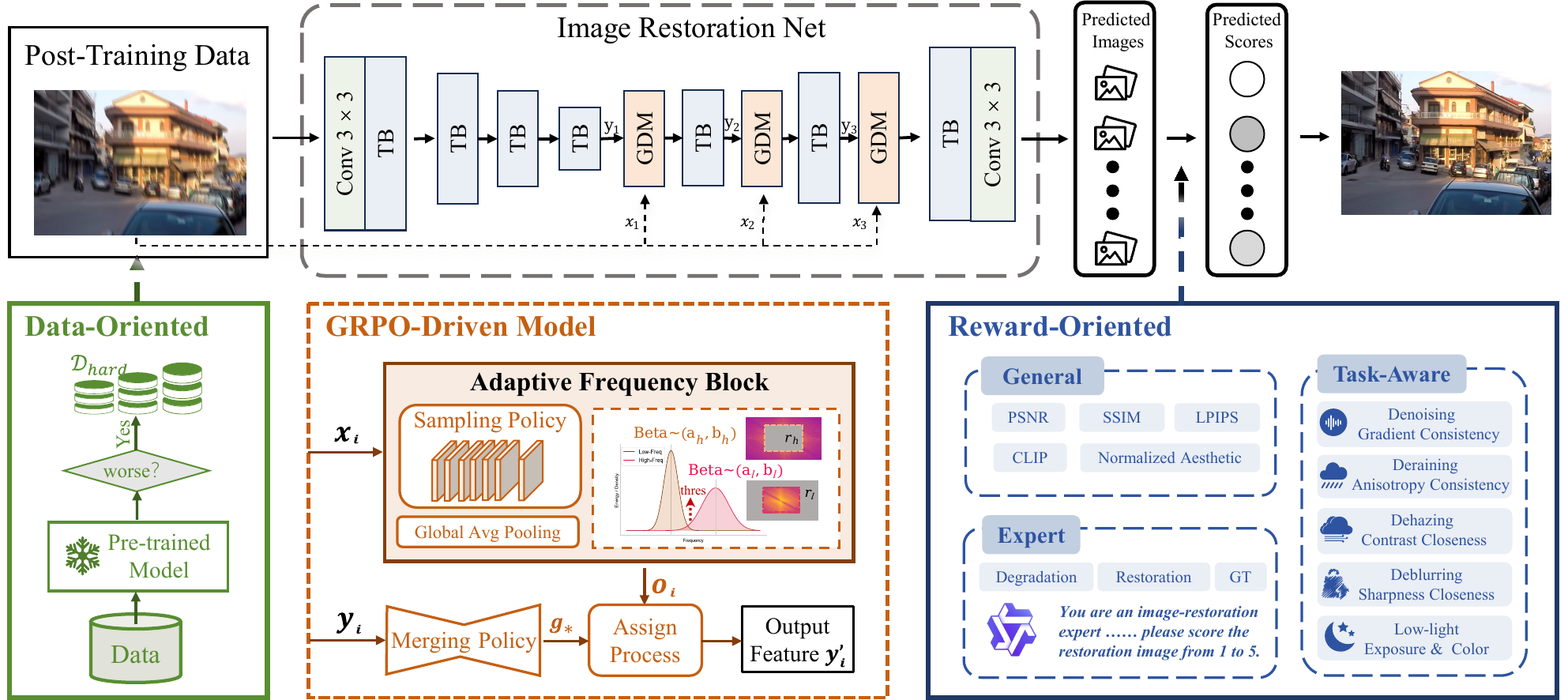}
    \vspace{-1mm}
    \caption{Overview of IRPO. Pillar 1 (Data-Oriented, left): a pre-trained model evaluates the full dataset and curates $\mathcal{D}_{\text{hard}}$ for post-training. Pillar 2 (Reward-Oriented, right): a multi-component reward model (General, Expert, Task-Aware) trains the policy $\pi_\theta$. Lightweight GRPO-Driven Model (GDM) heads adapt the restoration backbone inside the AFLB modules.}
    \label{fig:pipeline} 
    % \vspace{-0.5cm}
\end{figure*}
    % The policy itself is implemented as a GRPO-Driven Model (bottom-middle) that stochastically modulates the main Image Restoration Net (top).
\subsection{A Post-Training Paradigm for Low-Level Vision}
\label{sec:overview}

The prevailing training scheme for image restoration is end-to-end supervision. This approach has two main limitations. First, it treats all data points equally and may underfit complex, long-tail degradations. Second, it relies on deterministic pixel-level objectives, which can produce overly smooth results and gives limited feedback about degradation-specific artifacts. % metrics, an

We therefore add a focused post-training phase after the initial restoration model is trained. The goal is to improve robustness while preserving restoration fidelity. As illustrated in Fig.~\ref{fig:pipeline}, IRPO has two main pillars: Data-Oriented Supervision, which focuses on underperforming data (Sec.~\ref{sec:pillar1}), and Reward-Oriented Optimization, which incorporates quality-aware and task-aware feedback (Sec.~\ref{sec:pillar2}).

% The prevailing paradigm for training image restoration models is constrained by its reliance on pixel-level fidelity losses, such as $\ell_1$ or $\ell_2$. This optimization objective is often fundamentally misaligned with human perceptual preferences (e.g., texture, realism, aesthetic appeal) and typically leads to overly smooth reconstructions. Furthermore, this end-to-end supervision struggles to rectify complex degradations found in the long-tail of the data distribution, limiting the model's robustness.

% To transcend these limitations, we re-frame the post-training process. We introduce a novel Post-Training Paradigm for low-level vision, meticulously designed to systematically enhance both robustness and perceptual quality. As illustrated in Figure~\ref{fig:pipeline}, our paradigm is a dichotomy composed of two synergistic pillars: \textit{Data-Oriented Supervision}, which elevates the model's floor performance by rectifying failures on hard data (Sec.~\ref{sec:pillar1}), and \textit{Reward-Oriented Optimization}, which elevates the model's ceiling performance by optimizing for complex perceptual rewards (Sec.~\ref{sec:pillar2}).
\vspace{-0.1cm}
\subsection{Pillar 1: Data-Oriented Supervision}
\label{sec:pillar1}

\paragraph{Exploration on Data Curation.}
We first investigate how data selection affects post-training. Fine-tuning on the full training dataset, or on randomly sampled subsets, yields only limited gains (Fig.~\ref{fig:data_exploration_graph}). 
The results indicate that post-training is most useful when it focuses on the hard tail of the data distribution: samples where the pre-trained model performs poorly.

Based on this observation, we form a subset $\mathcal{D}_{\text{hard}} \subset \mathcal{D}$ from the worst-performing 30\% of the training data. The Data-Oriented pillar applies the post-training supervision to this subset.
% \vspace{-0.45cm}
\paragraph{Differentiable Supervision on $\mathcal{D}_{\text{hard}}$.}
For this hard-data subset, we optimize the restoration backbone $\mathcal{F}$ using two differentiable, data-driven losses. First, we identify the best-scoring sample ${y}_{g^\star}$ (where $g^\star=\arg\max_{i} r_i$ via our reward model) from a group of $G$ stochastic rollouts (see Sec.~\ref{sec:pillar2}). We then impose a standard $\ell_1$ loss:
\vspace{-0.1cm}
\begin{equation}
\mathcal{L}_{\mathrm{sup}}=\ell_1\big(y_{g^\star},\,\hat y\big),
\label{eq:sup_loss}
\end{equation}
where $\hat y$ is the ground-truth clean image. This provides a stable gradient signal to the backbone.

Second, to ensure stability, we compute a deterministic output ${y}_{\text{det}}$ (using the mean of the Beta policy distributions) and enforce a consistency objective:
\vspace{-0.1cm}
\begin{equation}
\mathcal{L}_{\mathrm{cons}}=\ell_1\big(y_{g^\star},\,{y}_{\text{det}}\big),
\label{eq:cons_loss}
\end{equation}
This loss anchors the stochastic policy's best output to its stable, deterministic counterpart.

\vspace{-0.1cm}
\subsection{Pillar 2: Reward-Oriented Optimization}
\label{sec:pillar2}

The supervised objectives in Pillar~1 primarily emphasize pixel-level fidelity (e.g., $\ell_1$), which can be insufficient to capture perceptual artifacts and degradation-specific residuals.
To incorporate richer feedback, we introduce reward terms that are important for restoration quality yet inconvenient (or impossible) to use as direct training losses, including $R_{\text{expert}}$ (an expert-judgment score) and $R_{\text{aes}}$ (an aesthetic score, as part of $R_{\text{gen}}$).
Since these signals are non-differentiable with respect to the restoration output, we treat post-training as policy optimization and employ GRPO to update the policy $\pi_\theta$ under the composite reward.

% The supervised losses in Pillar 1 can only optimize for pixel-level fidelity (L1). To align our model with complex human preferences, we must incorporate non-differentiable reward signals. Our reward function (detailed below) includes metrics crucial for perception but unusable as a direct loss, such as $R_{\text{qwen}}$ (worked as a expert judge) and $R_{\text{aes}}$ (aesthetic score, a component of $R_{\text{gen}}$). Since gradients cannot be backpropagated through these scores, we \emph{must} adopt a policy gradient method (GRPO) to optimize our policy $\pi_\theta$ using these rewards.
% \vspace{-0.45cm}
\paragraph{Reward Modeling.}
\label{sec:reward_model}
We design a composite reward signal with three components:
\vspace{-0.1cm}
\begin{equation}
\label{eq:reward_main}
R(y_{g^\star}, \hat y) = \lambda_{\text{gen}}R_{\text{gen}} + \lambda_{\text{expert}}R_{\text{expert}} + \lambda_{\text{task}}R_{\text{task}},
\end{equation}
where $y_{g^\star}$ is the restored output, $\hat y$ is the ground-truth, and $\lambda_{\text{gen}}, \lambda_{\text{expert}}, \lambda_{\text{task}}$ are fixed balancing hyperparameters.

\begin{itemize}
    \item \textbf{Generic Quality} ($R_{\text{gen}}$): A reward that combines LPIPS, CLIP similarity, and Aesthetic score with PSNR and SSIM to preserve fidelity while adding quality cues.
    \item \textbf{Expert Quality} ($R_{\text{expert}}$): a coarse (1--5) score produced by an expert visual judge. We treat this term as a high-level quality complement rather than a standalone low-level artifact detector.
    \item \textbf{Task-Aware Reward} ($R_{\text{task}}$): A specialized heuristic reward (\textit{e.g.,} gradient consistency for denoising) that guides the policy to restore the relevant attributes for each specific degradation task.
\end{itemize}
Appendix~\ref{app:reward_details} gives the detailed reward definitions.
% ~\ref{app:reward_details}
% \vspace{-0.45cm}
\paragraph{Policy Parametrization.}
\label{sec:policy_param}
Unlike generative methods that can sample diverse pixel-level outputs, restoration backbones such as AdaIR are deterministic. To apply GRPO, we adapt the backbone $\mathcal{F}$ into a stochastic policy $\pi_{\theta}$ by adding lightweight GRPO-Driven Model (GDM) heads to its AFLB modules. These heads do not generate images directly. They expose bounded continuous controls for frequency masks and feature fusion, so the policy acts as an input-adaptive router around the original restoration network. As shown in Fig.~\ref{fig:pipeline} (bottom center), the action $a=(r_h,r_l,g_f,g_o)$ contains two frequency-mask rates and two fusion weights. A policy head $\pi_{\theta}$ outputs Beta distribution parameters $(\alpha, \beta) \in \mathbb{R}_{+}^{4}$ for these controls:
% To be optimized by RL, we adapt the deterministic AdaIR~\cite{cui2025adair} backbone $\mathcal{F}$ into a stochastic policy $\pi_\theta$. We design a GRPO-Driven Model (GDM) adding lightweight policy heads to adapt to the stochastic characteristics of GRPO (see Figure~\ref{fig:pipeline}, bottom-middle), exposing four continuous controls $(r_h, r_l, g_f, g_o)$ as the action $a$. A policy head $\pi_\theta$ outputs Beta distribution parameters $(\boldsymbol{\alpha}, \boldsymbol{\beta}) \in \mathbb{R}_{+}^{4}$ for these controls:
\vspace{-0.1cm}
\begin{align}
a_k \sim \mathrm{Beta}(\alpha_k, \beta_k), \quad \text{for } k \in \{h, l, f, o\}.
\label{eq:beta_sampling}
\end{align}
The joint log-probability $\log \pi_\theta(a\!\mid\! x)$ has a closed form. The first two actions, $(r_h, r_l)_i$, control the frequency-domain mask used to extract high- and low-frequency components from input image $x_i$ and generate an intermediate feature $O_i$. The final two actions, $(g_f, g_o)_i$, dynamically fuse this feature with the original latent feature $y_i$:
\vspace{-0.1cm}
\begin{equation}
\label{eq:aflb_fusion}
y'_i = g_{f,i} \cdot y_i + g_{o,i} \cdot O_i,\quad \text{for } i \in \{1,2,3\},
\end{equation}
where $y'_i$ is the new feature passed to the next decoder block. This stochastic modulation produces the restored candidate ${y_g} = \mathcal{F}(x)$. We sample $G$ actions for each input image $x$. Appendix~\ref{app:policy_details} provides the policy details.
% ~\ref{app:policy_details}

% ~\ref{app:grpo_details}
\paragraph{Optimization Objective.}
We optimize the policy $\pi_\theta$ with GRPO under the non-differentiable reward $R$. Appendix~\ref{app:grpo_details} gives the GRPO background. GRPO constructs a \emph{group-normalized} relative advantage $A_i$ from a group of $G$ rollouts, and uses it in a clipped PPO-style surrogate objective. Our RL objective, $\mathcal{L}_{\mathrm{RL}}$, is:
\vspace{-0.1cm}
\begin{equation}
\label{eq:rl_loss}
\begin{aligned}
\mathcal{L}_{\mathrm{RL}}(\theta) = & -\,\frac{1}{G}\sum_{i=1}^{G} \min\Big(\rho_i(\theta)A_i, \clip(\rho_i(\theta), 1-\epsilon, 1+\epsilon)A_i\Big) \\
& +\, \beta\,\KLhat(\pi_\theta\|\pi_{\mathrm{ref}}) - \tau\,\Hent(\pi_\theta),
\end{aligned}
\end{equation}
% where $\rho_i(\theta)$ is the likelihood ratio, $A_i$ is the group-normalized advantage, $\beta$ controls the KL penalty, and $\tau$ controls the entropy bonus. 
% where $\rho_i(\theta)$ is the likelihood ratio, $A_i$ is the group-normalized advantage, $\beta$ controls the KL penalty, and $\tau$ is the entropy bonus. The reference policy $\pi_{\mathrm{ref}}$ used for the KL penalty.
% where $\rho_i(\theta)$ is the likelihood ratio, $A_i$ is the group-normalized advantage, $\beta$ controls the KL penalty, $\tau$ is the entropy bonus, and $\pi_{\mathrm{ref}}$ is the reference policy used for KL regularization and $\pi_\theta$ is the policy being optimized.
where $\rho_i(\theta)$ is the likelihood ratio, $A_i$ is the group-normalized advantage, $\beta$ controls the KL penalty, $\tau$ is the entropy bonus, $\pi_{\mathrm{ref}}$ denotes the reference policy for KL regularization, and $\pi_\theta$ is the policy being optimized.

\subsection{Final Hybrid Objective}
\label{sec:final_objective}
Our final objective combines the pillars. We jointly optimize the policy $\pi_\theta$ (via $\mathcal{L}_{\mathrm{RL}}$ from Pillar 2) and the backbone $\mathcal{F}$ (via $\mathcal{L}_{\mathrm{sup}}$ and $\mathcal{L}_{\mathrm{cons}}$ from Pillar 1). All computations are performed on our curated hard-data subset $\mathcal{D}_{\text{hard}}$:
\vspace{-0.1cm}
\begin{equation}
\mathcal{L}_{\mathrm{total}}(\theta)
= \mathbb{E}_{(x,g)\sim\mathcal{D}_{\text{hard}}}
\big[\,\mathcal{L}_{\mathrm{RL}}(\theta)
+ \lambda_{\mathrm{sup}}\mathcal{L}_{\mathrm{sup}}
+ \lambda_{\mathrm{cons}}\mathcal{L}_{\mathrm{cons}}
\,\big],
\label{eq:total_loss}
\end{equation}
where $\lambda_{\mathrm{sup}}$ and $\lambda_{\mathrm{cons}}$ are weights with linear annealing schedules (see Appendix~\ref{app:imp_details}).

\section{Experiment}
\label{sec:experiment}

\subsection{Experimental Setup}
\label{sec:exp_setup}
\noindent\textbf{Datasets.}\label{sec:datasets}
For training and testing, we follow the standard protocols used in prior all-in-one restoration works~\citep{promptir, airnet, cui2025adair}.
\vspace{-0.15cm}
\paragraph{In-Domain (ID) Datasets.}
Our training and evaluation cover five restoration tasks: dehazing on SOTS~\citep{RESIDE}, deraining on Rain100L~\citep{rain100L}, deblurring on GoPro~\citep{gopro}, low-light enhancement on LOL~\citep{lol}, and Gaussian denoising trained on BSD400~\citep{bsd400}+WED~\citep{wed} and evaluated on BSD68~\citep{bsd68}/Urban100~\citep{urban100} with $\sigma\in\{15,25,50\}$.
%
% Finally, under the all-in-one setting, we train a single model on the combined set of the aforementioned training datasets, and directly test it across multiple restoration tasks (as shown in Table~\ref{tab:5D} and Table~\ref{tab:3D}).
\vspace{-0.15cm}

\begin{table*}[t]
\centering
\caption{Comparisons for five-degradation all-in-one restoration. Each cell reports PSNR/SSIM, and denoising is evaluated at $\sigma=25$. All-in-one baselines are reported under the same five-task protocol; IRPO improves the AdaIR backbone by 0.93~dB average PSNR.}
\label{tab:5D}
\maintablefont
\begin{adjustbox}{max width=\textwidth}
\begin{tabular}{@{}lcccccc@{}}
\toprule
\tablehead
Method & \makecell{Dehaze\\SOTS} & \makecell{Derain\\Rain100L} & \makecell{Denoise\\CBSD68} & \makecell{Deblur\\GoPro} & \makecell{Low-Light\\LOL} & \makecell{Average\\PSNR/SSIM} \\
\midrule
NAFNet~\citep{chen2022simple} & 25.23/0.939 & 35.56/0.967 & 31.02/0.883 & 26.53/0.808 & 20.49/0.809 & 27.76/0.881 \\
HINet~\citep{hinet} & 24.74/0.937 & 35.67/0.969 & 31.00/0.881 & 26.12/0.788 & 19.47/0.800 & 27.40/0.875 \\
MPRNet~\citep{MPRNet} & 24.27/0.937 & 38.16/0.981 & 31.35/0.889 & 26.87/0.823 & 20.84/0.824 & 28.30/0.891 \\
DGUNet~\citep{dgunet} & 24.78/0.940 & 36.62/0.971 & 31.10/0.883 & 27.25/0.837 & 21.87/0.823 & 28.32/0.891 \\
MIRNet-v2~\citep{mirnet} & 24.03/0.927 & 33.89/0.954 & 30.97/0.881 & 26.30/0.799 & 21.52/0.815 & 27.34/0.875 \\
SwinIR~\citep{SwinIR} & 21.50/0.891 & 30.78/0.923 & 30.59/0.868 & 24.52/0.773 & 17.81/0.723 & 25.04/0.835 \\
Restormer~\citep{restormer} & 24.09/0.927 & 34.81/0.962 & 31.49/0.884 & 27.22/0.829 & 20.41/0.806 & 27.60/0.881 \\
\midrule
DL~\citep{dl} & 20.54/0.826 & 21.96/0.762 & 23.09/0.745 & 19.86/0.672 & 19.83/0.712 & 21.05/0.743 \\
Transweather~\citep{transweather} & 21.32/0.885 & 29.43/0.905 & 29.00/0.841 & 25.12/0.757 & 21.21/0.792 & 25.22/0.836 \\
TAPE~\citep{tape} & 22.16/0.861 & 29.67/0.904 & 30.18/0.855 & 24.47/0.763 & 18.97/0.621 & 25.09/0.801 \\
AirNet~\citep{airnet} & 21.04/0.884 & 32.98/0.951 & 30.91/0.882 & 24.35/0.781 & 18.18/0.735 & 25.49/0.846 \\
IDR~\citep{idr} & 25.24/0.943 & 35.63/0.965 & 31.60/0.887 & 27.87/0.846 & 21.34/0.826 & 28.34/0.893 \\
GridFormer~\citep{gridformer} & 26.79/0.951 & 36.61/0.971 & 31.45/0.885 & 29.22/0.884 & 22.59/0.831 & 29.33/0.904 \\
InstructIR~\citep{instructir} & 27.10/0.956 & 36.84/0.973 & 31.40/0.887 & 29.40/0.886 & 23.00/0.836 & 29.55/0.907 \\
VLU-Net~\citep{vlunet} & 30.84/0.980 & 38.54/0.982 & 31.43/0.891 & 27.46/0.840 & 22.29/0.833 & 30.11/0.905 \\
DFPIR~\citep{dfpir} & 31.64/0.979 & 37.62/0.978 & 31.29/0.889 & 28.82/0.873 & 23.82/0.843 & 30.64/0.913 \\
\alternativerow AdaIR~\citep{cui2025adair} & 30.53/0.978 & 38.02/0.981 & 31.35/0.889 & 28.12/0.858 & 23.00/0.845 & 30.20/0.910 \\
\methodrow \textbf{IRPO (Ours)} & \textbf{31.85/0.984} & \textbf{38.72/0.987} & \textbf{32.13/0.896} & \textbf{28.89/0.863} & \textbf{24.08/0.858} & \textbf{31.13/0.918} \\
\bottomrule
\end{tabular}
\end{adjustbox}
\vspace{-4pt}
\end{table*}

\paragraph{Out-of-Domain (OOD) Datasets.}
To evaluate generalization, we test the same all-in-one model on five OOD datasets: RRSHID~\citep{zhu2025real} for real dehazing, GT-Rain~\citep{ba2022not} for real deraining, SIDD~\citep{abdelhamed2018high} for smartphone denoising, ReLoBlur~\citep{li2023real} for local-motion deblurring, and LoLi-Street~\citep{islam2024loli} for street-scene low-light enhancement.

% To validate the efficacy of the proposed IRPO, we conduct experiments under two different settings:
% (1) All-in-One, and (2) Out-of-Domain. In the All-in-One setting, a unified model is trained to perform image restoration across multiple degradation types.
% We provide additional ablation experiments, visual examples,
% and more details on the architecture in the supplementary material.
\vspace{-0.15cm}

\begin{table*}[t]
\centering
\caption{OOD real-world evaluation with the same AdaIR backbone. Each cell reports PSNR/SSIM, and the average is computed over all five unseen degradation sources.}
\label{tab:five_deg_metrics}
\maintablefont
\begin{adjustbox}{max width=\textwidth}
\begin{tabular}{@{}lcccccc@{}}
\toprule
\tablehead
Method & \makecell{RRSHID\\Dehaze} & \makecell{GT-Rain\\Derain} & \makecell{SIDD\\Denoise} & \makecell{ReLoBlur\\Deblur} & \makecell{LoLi-Street\\Low-Light} & \makecell{Average\\PSNR/SSIM} \\
\midrule
\alternativerow AdaIR & 16.71/0.475 & 19.58/0.684 & 28.83/0.902 & 20.43/0.806 & 15.96/0.781 & 20.30/0.730 \\
\methodrow \textbf{IRPO} & \textbf{19.94/0.623} & \textbf{21.17/0.692} & \textbf{33.18/0.919} & \textbf{21.94/0.837} & \textbf{22.43/0.875} & \textbf{23.73/0.789} \\
\bottomrule
\end{tabular}
\end{adjustbox}
\vspace{-4pt}
\end{table*}

\paragraph{Implementation Details.} IRPO is initialized from AdaIR~\citep{cui2025adair}. It keeps the 4-level encoder-decoder and three AFLB modules, and adds lightweight policy heads for the Beta actions $(r_h,r_l,g_f,g_o)$. We fine-tune the worst-performing 30\% training samples for 30 epochs with group size $G=4$. We use Adam with learning rate $3\times10^{-5}$ and a $6\times$ learning-rate multiplier for the policy heads. We keep AdaIR's $128\times128$ crops and augmentations, and add annealed $\mathcal{L}_1$ and consistency losses. Training uses two NVIDIA 4090 GPUs plus two GPUs serving Qwen2.5-VL-7B-Instruct~\citep{qwen25vl}. At inference, the VLM is removed and the Beta policy uses deterministic expectations, so IRPO has no extra VLM overhead. Appendix~\ref{app:imp_details} lists the full hyperparameters.
 % (Table~\ref{tab:hyperparameters})
\subsection{Main Experiments}
\label{sec:main_exp}

We evaluate IRPO on standard all-in-one restoration settings and single-task benchmarks.

\begin{figure*}[t!]
    \centering
    \includegraphics[width=0.96\textwidth]{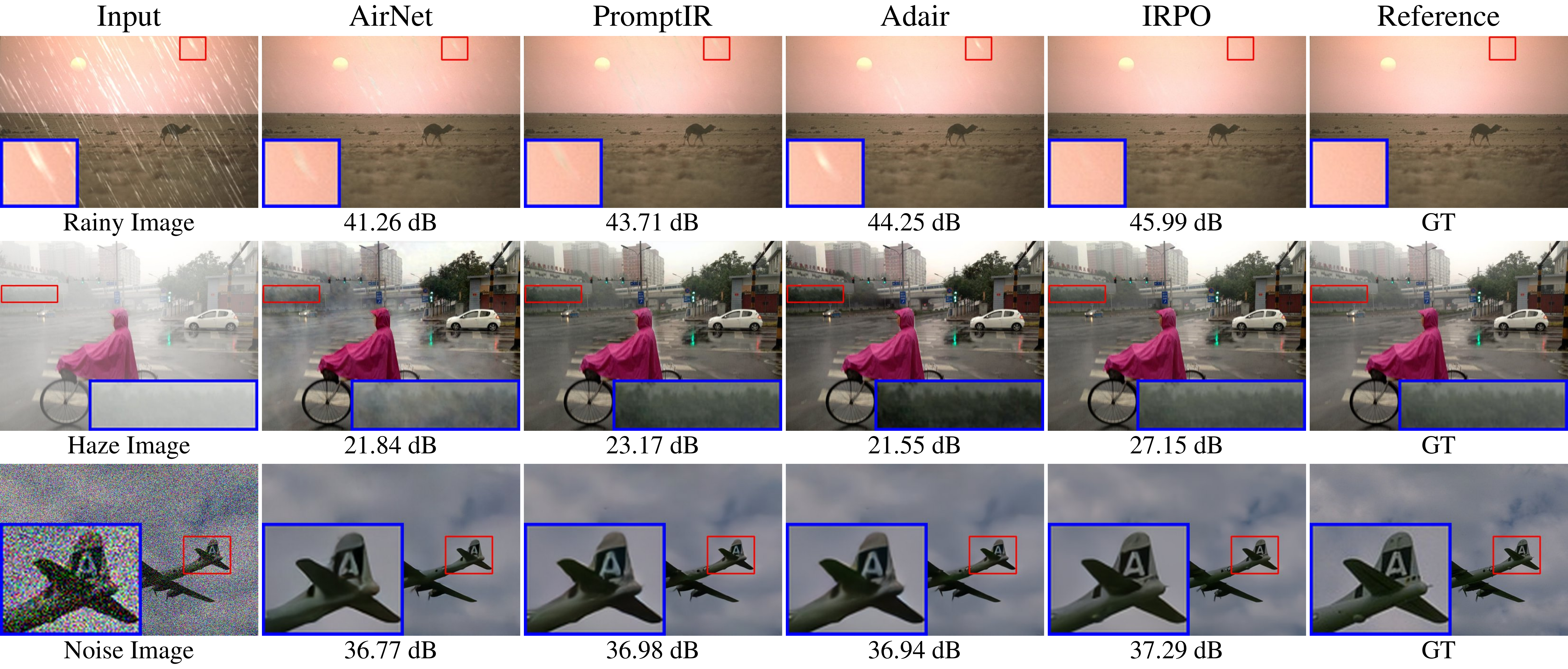}
    \vspace{-2mm}
    \caption{Qualitative comparisons on synthetic restoration benchmarks for deraining, dehazing, and denoising. IRPO removes residual degradations more effectively than AirNet, PromptIR, and AdaIR while preserving local structures and fine details.}
    \label{fig:synthetic_visual}
    \vspace{-7pt}
\end{figure*}

\paragraph{All-in-One Restoration.}
Table~\ref{tab:5D} compares IRPO with general IR methods and all-in-one restorers, including 2024--2025 baselines under the same five-task protocol. IRPO obtains the best average performance and improves AdaIR by 0.93~dB average PSNR, with gains on dehazing (+1.32~dB), low-light enhancement (+1.08~dB), and denoising (+0.78~dB). Table~\ref{tab:3D} further evaluates the three-degradation setting with haze, rain, and noise. IRPO remains the top performer and improves AdaIR by 0.74~dB average PSNR, suggesting that the post-training gain is consistent across both compact and broader all-in-one protocols. Fig.~\ref{fig:synthetic_visual} provides synthetic visual comparisons; additional per-task and real-world visual examples are included in Appendix~\ref{app:visual_results}.

\vspace{-0.15cm}
\paragraph{Single-Task Restoration.}
Appendix Table~\ref{tab:panel_three} reports specialized single-task models. IRPO consistently improves over AdaIR and PromptIR across dehazing, deraining, and denoising, with the largest gain on Urban100 ($\sigma=50$, +0.37~dB).

% ~\ref{app:additional_results}
% ~\ref{app:additional_visual}

\begin{table*}[t]
\centering
\caption{Comparisons under the three-degradation all-in-one setting. Each cell reports PSNR/SSIM; denoising is evaluated on CBSD68 at three noise levels. IRPO improves the AdaIR backbone by 0.74~dB average PSNR.}
\label{tab:3D}
\footnotesize\setlength{\tabcolsep}{2.4pt}\renewcommand{\arraystretch}{0.90}
\begin{adjustbox}{max width=\textwidth}
\begin{tabular}{@{}lcccccc@{}}
\toprule
\tablehead
Method & \makecell{Dehaze\\SOTS} & \makecell{Derain\\Rain100L} & \multicolumn{3}{c}{Denoising on CBSD68} & Average \\
\cmidrule(lr){4-6}
\tablehead
 & PSNR/SSIM & PSNR/SSIM & $\sigma=15$ & $\sigma=25$ & $\sigma=50$ & PSNR/SSIM \\ 
\midrule
BRDNet~\citep{brdnet} & 23.23/0.895 & 27.42/0.895 & 32.26/0.898 & 29.76/0.836 & 26.34/0.693 & 27.80/0.843 \\
MPRNet~\citep{MPRNet} & 25.28/0.955 & 33.57/0.954 & 33.54/0.927 & 30.89/0.880 & 27.56/0.779 & 30.17/0.899 \\
AirNet~\citep{airnet} & 27.94/0.962 & 34.90/0.968 & 33.92/0.933 & 31.26/0.888 & 28.00/0.797 & 31.20/0.910 \\
PromptIR~\citep{promptir} & 30.58/0.974 & 36.37/0.972 & 33.98/0.933 & 31.31/0.888 & 28.06/0.799 & 32.06/0.913 \\
GridFormer~\citep{gridformer} & 30.37/0.970 & 37.15/0.972 & 33.93/0.931 & 31.37/0.887 & 28.11/0.801 & 32.19/0.912 \\
InstructIR~\citep{instructir} & 30.22/0.959 & 37.98/0.978 & 34.15/0.933 & 31.52/0.890 & 28.30/0.804 & 32.43/0.913 \\
VLU-Net~\citep{vlunet} & 30.71/0.980 & 38.93/0.984 & 34.13/0.935 & 31.48/0.892 & 28.23/0.804 & 32.70/0.919 \\
DFPIR~\citep{dfpir} & 31.87/0.980 & 38.65/0.982 & 34.14/0.935 & 31.47/0.893 & 28.25/0.806 & 32.88/0.919 \\
\alternativerow AdaIR~\citep{cui2025adair} & 31.06/0.980 & 38.64/0.983 & 34.12/0.935 & 31.45/0.892 & 28.19/0.802 & 32.69/0.918 \\
\methodrow \textbf{IRPO (Ours)} & \textbf{31.88/0.984} & \textbf{39.13/0.988} & \textbf{34.54/0.938} & \textbf{32.04/0.905} & \textbf{29.56/0.823} & \textbf{33.43/0.928} \\
\bottomrule
\end{tabular}
\end{adjustbox}
\vspace{-4pt}
\end{table*}

\begin{table*}[t]
\centering
\caption{Attribution of reward components under the same hard-data subset and GRPO schedule. Each cell reports PSNR/SSIM; the final column reports the average PSNR gain over AdaIR.}
\label{tab:five_deg_psnr_ssim_combo}
\maintablefont
\begin{adjustbox}{max width=\textwidth}
\begin{tabular}{@{}lccccccc@{}}
\toprule
\tablehead
Method & Dehazing & Deraining & Denoising & Deblurring & Low-Light & \makecell{Average\\PSNR/SSIM} & \makecell{$\Delta$Avg.\\PSNR} \\
\midrule
\alternativerow AdaIR & 30.53/0.978 & 38.02/0.981 & 31.35/0.889 & 28.12/0.858 & 23.00/0.845 & 30.20/0.910 & 0.00 \\
+ $R_{\text{gen}}$ & 31.30/0.981 & 38.57/0.985 & 31.95/0.893 & 28.64/0.861 & 23.68/0.852 & 30.83/0.914 & +0.63 \\
+ $R_{\text{expert}}$ & 30.90/0.980 & 38.28/0.983 & 31.64/0.891 & 28.35/0.859 & 23.31/0.848 & 30.50/0.912 & +0.30 \\
+ $R_{\text{task}}$ & 31.12/0.981 & 38.42/0.984 & 31.83/0.893 & 28.57/0.861 & 23.57/0.850 & 30.70/0.914 & +0.50 \\
\methodrow \textbf{IRPO} & \textbf{31.85/0.984} & \textbf{38.72/0.987} & \textbf{32.13/0.896} & \textbf{28.89/0.863} & \textbf{24.08/0.858} & \textbf{31.13/0.918} & \textbf{+0.93} \\
\bottomrule
\end{tabular}
\end{adjustbox}
\vspace{-4pt}
\end{table*}

\begin{figure}[t!]
    \centering
    \includegraphics[width=0.62\linewidth]
    {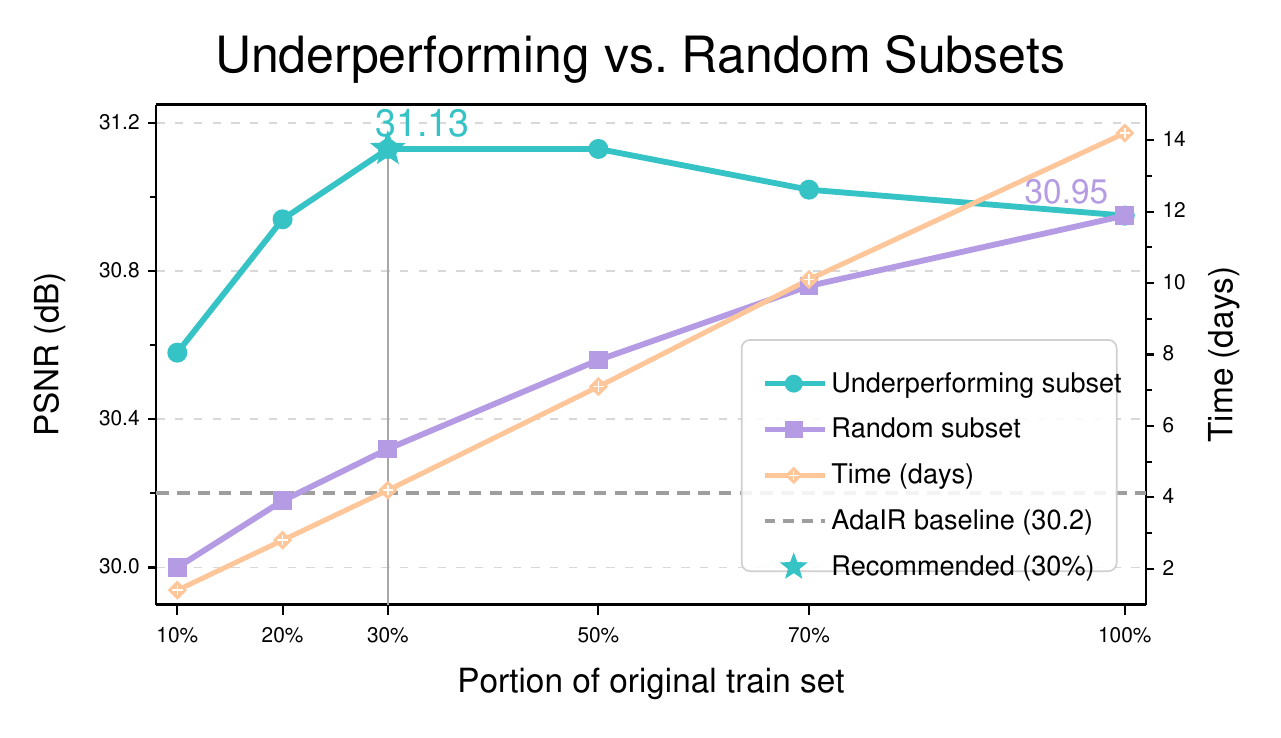}
    % \vspace{-7mm}
    \caption{Post-training on underperforming \textit{vs.} random subsets for the five-task all-in-one setting. The left y-axis shows average PSNR and the right y-axis shows time. The 30\% underperforming subset reaches 31.13~dB in about 4.2 GPU days; 100\% data reaches 30.95~dB with roughly three times the cost.}
    \label{fig:data_exploration_graph} 
\vspace{-4pt}
\end{figure}

\subsection{Evaluation on OOD Generalization}
\label{sec:exp_ood}

OOD generalization is central for all-in-one restoration, so we evaluate IRPO on unseen degradation sources in Table~\ref{tab:five_deg_metrics}. IRPO raises AdaIR from 20.30/0.730 to 23.73/0.789 average PSNR/SSIM, with the largest gains on low-light (+6.47~dB) and denoising (+4.35~dB). These results support the main design of IRPO: hard-case post-training with quality-aware rewards improves OOD robustness while retaining supervised fidelity constraints. Real-world visual examples in Appendix~\ref{app:visual_results} show the same trend; expert-source, group-size, efficiency, training-curve, data-strategy, and reward-weight analyses are in Appendix~\ref{app:additional_analysis} for completeness.

% These results demonstrate that our model exhibits strong generalization ability in OOD real-world scenarios, accurately identifying unseen image degradations and effectively restoring them to realistic, high-quality images.
\vspace{-0.2cm}
\subsection{Ablation Study}
\label{sec:ablation}

We focus on data selection and reward composition in the main paper, and report group-size, expert-source, efficiency, data-ratio, reward-weight, and stability analyses in Appendix~\ref{app:additional_analysis}.
\vspace{-0.15cm}
\paragraph{Effect of Data Portion.}
Fig.~\ref{fig:data_exploration_graph} shows that the 30\% underperforming subset reaches 31.13~dB in about 4.2 GPU days, while 100\% data reaches 30.95~dB with roughly three times the cost. Appendix~\ref{app:additional_analysis} further analyzes the selected feature space, uniform \textit{vs.} adaptive data ratios, and reward-weight sensitivity.
% \vspace{-0.4cm}
\vspace{-0.2cm}
\paragraph{Effect of Reward Components.}
Table~\ref{tab:five_deg_psnr_ssim_combo} isolates reward components. $R_{\text{gen}}$ gives the largest individual gain (+0.63~dB) because it keeps full-reference fidelity signals. $R_{\text{task}}$ (+0.50~dB) targets degradation residuals, and $R_{\text{expert}}$ (+0.30~dB) acts as a coarse quality complement. Full IRPO reaches 31.13~dB, outperforming every single-reward variant.

\paragraph{Contribution of data and rewards.}
Fig.~\ref{fig:data_exploration_graph} changes the post-training data, while Table~\ref{tab:five_deg_psnr_ssim_combo} fixes the hard subset and GRPO update to isolate reward channels. Thus, the gains come from hard-tail data selection plus reward-guided policy optimization, not from any single perceptual or VLM reward source alone or from score scaling.

\paragraph{Interaction with fidelity losses.}
The reward design does not replace supervised fidelity: GRPO selects among stochastic candidates on the hard subset, while $\mathcal{L}_{\mathrm{sup}}$ and $\mathcal{L}_{\mathrm{cons}}$ anchor updates to paired ground truth and deterministic behavior. Auxiliary quality rewards can therefore guide exploration without discarding distortion-oriented constraints during post-training.

\subsection{Post-Training Analysis}
\label{sec:further_analysis}

Table~\ref{tab:attribution_iqa_main}(a) separates the sources of improvement in IRPO. Complementary to Table~\ref{tab:five_deg_psnr_ssim_combo}, it changes data, heads, reward access, and update rules rather than individual reward channels. Fine-tuning AdaIR on the hard subset improves the five-degradation average but not OOD robustness, showing that hard-tail data are useful but insufficient. Adding stochastic heads with supervised losses brings a larger OOD gain. Reward-ranked distillation uses the same sampled candidates and reward model as IRPO, but remains below full GRPO optimization, indicating that group-relative policy learning contributes beyond reward-based candidate selection.

We also evaluate quality-oriented metrics beyond PSNR/SSIM in Table~\ref{tab:attribution_iqa_main}(b). Full-reference LPIPS~\citep{zhang2018unreasonable} and DISTS~\citep{dists} are computed on paired OOD benchmarks. NIQE~\citep{niqe} and MUSIQ~\citep{musiq} are computed on the real-world visual set without references. IRPO improves these perceptual/IQA scores over AdaIR and DFPIR, which is consistent with the synthetic comparisons in Fig.~\ref{fig:synthetic_visual} and the real-world visual examples in Appendix~\ref{app:visual_results} for qualitative inspection. These metrics are used as supporting evidence for quality-aware restoration, while the main objective remains fidelity-oriented restoration with data-driven and reward-driven post-training for robust restoration.

\begin{table}[t]
\centering
\caption{Additional analysis of IRPO. (a) Attribution study under the same AdaIR initialization; both averages report PSNR. (b) Quality-oriented evaluation with full-reference OOD metrics and no-reference real-world IQA metrics.}
\label{tab:attribution_iqa_main}
\maintablefont
\begin{adjustbox}{max width=\textwidth}
\begin{minipage}{0.62\textwidth}
\centering
\textbf{(a) Attribution of post-training gains}\\[-1pt]
\begin{adjustbox}{max width=\linewidth}
\begin{tabular}{@{}lcccccc@{}}
\toprule
\tablehead
Variant & Hard & Heads & Rew. & Rule & 5D & OOD \\
\midrule
AdaIR & -- & -- & -- & -- & 30.20 & 20.30 \\
Hard FT & \checkmark & -- & -- & Sup. & 30.43 & 20.22 \\
Sup. GDM & \checkmark & \checkmark & -- & Sup. & 30.75 & 22.16 \\
Reward distill. & \checkmark & \checkmark & \checkmark & Distill. & 30.91 & 22.89 \\
\methodrow IRPO-GRPO & \checkmark & \checkmark & \checkmark & GRPO & \textbf{31.13} & \textbf{23.73} \\
\bottomrule
\end{tabular}
\end{adjustbox}
\end{minipage}
\hspace{4pt}
\begin{minipage}{0.35\textwidth}
\centering
\textbf{(b) Perceptual/IQA evaluation}\\[-1pt]
\begin{adjustbox}{max width=\linewidth}
\begin{tabular}{@{}lcccc@{}}
\toprule
\tablehead
Method & LPIPS$\downarrow$ & DISTS$\downarrow$ & NIQE$\downarrow$ & MUSIQ$\uparrow$ \\
\midrule
AdaIR & 0.191 & 0.140 & 4.18 & 61.5 \\
DFPIR & 0.184 & 0.136 & 4.12 & 62.2 \\
\methodrow IRPO & \textbf{0.174} & \textbf{0.126} & \textbf{3.89} & \textbf{63.8} \\
\bottomrule
\end{tabular}
\end{adjustbox}
\end{minipage}
\end{adjustbox}
\vspace{-4pt}
\end{table}

\paragraph{Reading the controls.}
The attribution variants are deliberately matched in AdaIR initialization and then progressively add hard data, stochastic heads, reward access, and the update rule. This keeps the comparison focused on the source of improvement rather than on a hidden change in training data or model capacity. Reward-ranked distillation already benefits from the same sampled candidates and rewards as IRPO, but it collapses each group to a single pseudo-target; IRPO instead uses group-relative advantages to update the stochastic heads over frequency masks and feature-fusion gates. The IQA block should be read in the same spirit: LPIPS, DISTS, NIQE, and MUSIQ are diagnostics for whether the quality-aware rewards hurt visual naturalness, while PSNR/SSIM remain the primary restoration protocol.

\paragraph{Practical overhead.}
Appendix Table~\ref{tab:efficiency_main} summarizes the computational cost. IRPO adds only lightweight policy heads to AdaIR (+0.05M parameters), and the VLM reward is used only during post-training. At inference, the Beta policy is evaluated by its deterministic expectation and the Qwen-VL scorer is removed. The deployed model therefore keeps the same 1080p latency as AdaIR while improving the OOD average by 3.43~dB. The extra computation is a one-time post-training cost.

\paragraph{Additional diagnostics.}
The appendix first provides the experimental evidence behind the main claims. Appendix~\ref{app:additional_analysis} reports training curves, hard-sample visualization, single-task results, group-size and expert-source ablations, efficiency accounting, and reward-weight sensitivity. Appendix~\ref{app:visual_results} provides real-world and additional per-task visual comparisons. The remaining sections make the recipe auditable: Appendix~\ref{app:imp_details} lists hyperparameters, Appendix~\ref{app:grpo_details} derives the GRPO objective, Appendix~\ref{app:reward_details} expands the reward definitions, and Appendix~\ref{app:policy_details} gives the Beta-policy log probability.

\paragraph{Scope of the post-training wrapper.}
IRPO is a post-training recipe rather than a new restoration backbone. We instantiate it on AdaIR because its AFLB modules provide a natural place to attach stochastic policy heads, but the data-selection rule and reward model operate at the dataset and output-image levels. They only require a pre-trained restorer, a hard-sample subset, and a train-time mechanism for generating multiple candidate restorations. Thus, AdaIR determines the specific action parameterization used here, while the broader recipe can be adapted to other encoder-decoder or transformer restorers with architecture-specific stochastic heads. At deployment, the policy is evaluated deterministically and the model keeps the same input-output interface as the backbone. Appendix~\ref{app:promptir_transfer} reports an additional evaluation on PromptIR.

\section{Conclusion}
This study introduces IRPO, a GRPO-based post-training framework for deterministic restoration models. IRPO has two core components. First, the data-oriented component focuses post-training on the 30\% underperforming samples, which improves the accuracy-efficiency trade-off. Second, the reward-oriented component combines fidelity, coarse visual-quality feedback, and task-specific restoration cues. Experiments on in-domain and OOD benchmarks show that IRPO improves the AdaIR backbone and strengthens real-world generalization. These results indicate that data-driven and reward-driven post-training is a practical direction for robust image restoration.

% ---- Bibliography ----
%
% BibTeX users should specify bibliography style 'plainnat' for NeurIPS.
% The NeurIPS style loads natbib automatically.
%
\renewcommand{\acksection}{\section*{Acknowledgments}}
\begin{ack}
This work was supported in part by the Natural Science Foundation of China under Grant 62403401, and in part by the Guangdong Basic and Applied Basic Research Foundation under Grant 2024A1515011992 and Grant 2026A1515012323.
This work was supported by the National Natural Science Foundation of China under Grant No. 62320106007, and by NUS Grant A-0010106-00-00, A-8004365-00-00 and A-8004410-01-00.
\end{ack}

\bibliographystyle{plainnat}
\bibliography{main}

% % 正文内容结束
\clearpage
\appendix
\raggedbottom
\numberwithin{equation}{section}
\numberwithin{figure}{section}
\numberwithin{table}{section}

\section{Additional Experimental Results and Analysis}
\label{app:additional_analysis}
\suppressfloats[t]

This section reports additional results and analyses. We first show training stability and hard-sample structure, then provide single-task evaluation, group-size and expert-source ablations, efficiency analysis, and reward-weight sensitivity.

\subsection{Training and Data-Selection Diagnostics}
\begin{figure}[!htbp]
    \centering
    \includegraphics[width=0.9\linewidth]{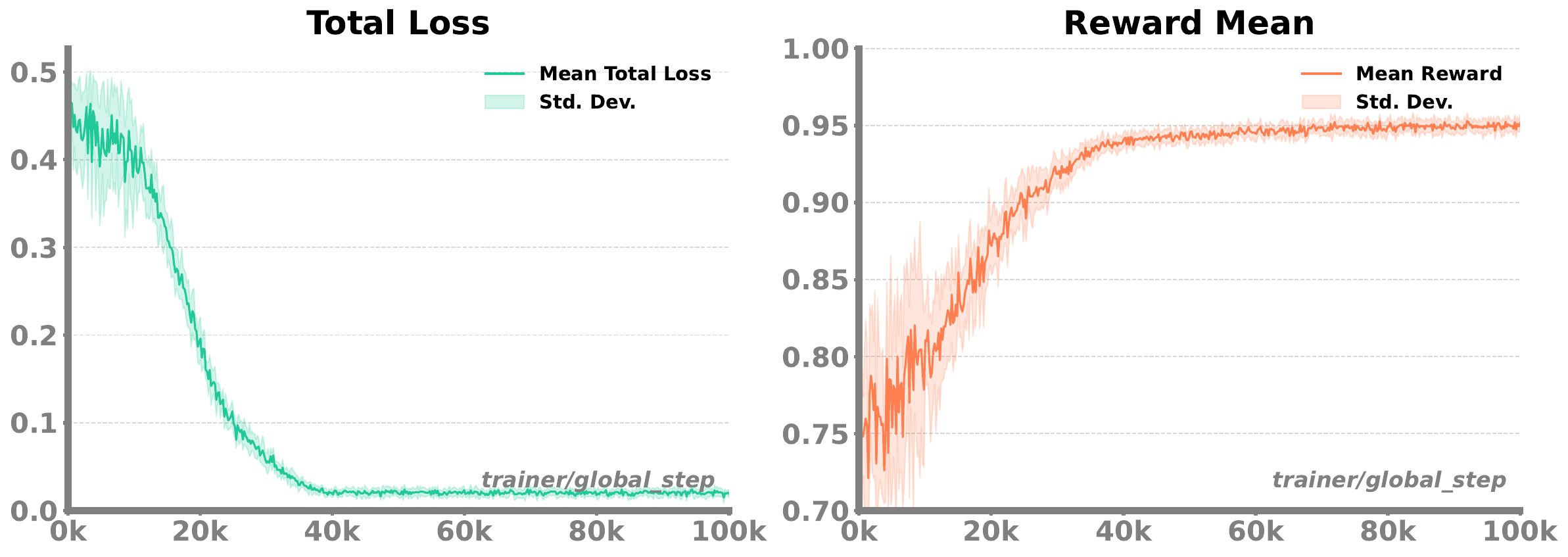}
    \caption{Training curves for IRPO post-training (\textit{mean} and \textit{std. dev.} over 30 epochs). The total loss smoothly converges, while the mean reward steadily increases and reaches a stable plateau.}
    \label{fig:training_curves}
\end{figure}

Figure~\ref{fig:training_curves} shows stable post-training dynamics. The total loss decreases smoothly, while the mean reward increases and reaches a plateau, indicating that the reward-guided stage remains numerically stable rather than oscillatory.

\begin{figure}[!htbp]
    \centering
    \includegraphics[width=0.45\linewidth]{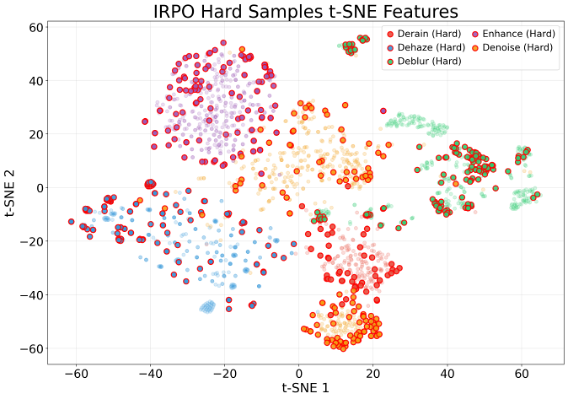}
    \caption{T-SNE feature visualization of the hard samples ($\mathcal{D}_{hard}$) across five tasks compared to the full dataset. The selected samples naturally cluster in complex, long-tail regions.}
    \label{fig:tsne_app}
\end{figure}

In Pillar 1 (Data-Oriented Supervision), fine-tuning on the 30\% underperforming samples gives the selected accuracy-efficiency trade-off. To examine this subset, we visualize the pre-trained feature space using T-SNE in Figure~\ref{fig:tsne_app}. The selected 30\% samples mainly lie in long-tail, high-entropy regions of the distribution. This indicates that the subset contains difficult cases, such as severe noise and complex occlusion, where the deterministic baseline performs less reliably.

The main experiments use a uniform 30\% ratio across degradation tasks. We also test an adaptive task-specific ratio, such as assigning 40\% to low-light enhancement and 20\% to deraining. As shown in Table~\ref{tab:data_strategy_app}, adaptive sampling slightly improves deblurring and low-light enhancement, but the uniform 30\% strategy gives the best average performance (31.13 dB).

\begin{table}[!htbp]
    \centering
    \caption{Comparison between uniform and adaptive data selection strategies. Performance is measured by average PSNR/SSIM on the all-in-one benchmark.}
    \label{tab:data_strategy_app}
    \appendixtablefont
    \begin{adjustbox}{max width=\columnwidth}
    \begin{tabular}{@{}llcccccc@{}}
    \toprule
    \tablehead
    Strategy & Ratio Config & Dehaze & Derain & Denoise & Deblur & Low-Light & Average \\
    \midrule
    \methodrow Uniform & All 30\% & \textbf{31.85/0.984} & \textbf{38.72/0.987} & \textbf{32.13/0.896} & 28.89/0.863 & 24.08/0.858 & \textbf{31.13/0.918} \\
    Adaptive & 30/20/25/35/40\% & 31.58/0.983 & 38.55/0.984 & 31.98/0.893 & \textbf{29.02/0.865} & \textbf{24.21/0.860} & 31.07/0.917 \\
    \bottomrule
    \end{tabular}
    \end{adjustbox}
\end{table}

\subsection{Single-Task Restoration Results}
Table~\ref{tab:panel_three} reports the single-task setting where each model is trained separately on its respective dataset. IRPO consistently improves over the AdaIR backbone across dehazing, deraining, and denoising, indicating that the post-training recipe is not limited to the mixed all-in-one setting.

\begin{table*}[t]
\centering
\caption{Single-task comparisons for dehazing (a), deraining (b), and denoising (c). For this setting, each model is trained separately on its respective dataset. All metrics shown are PSNR/SSIM.}
\label{tab:panel_three}
\appendixwidetablefont
\begin{adjustbox}{max width=\textwidth}
\begin{tabular}{@{}lc lc l ccc ccc c@{}}
\toprule
\tablehead
\multicolumn{2}{c}{(a) Dehaze (SOTS)} & \multicolumn{2}{c}{(b) Derain (Rain100L)} & \multicolumn{8}{c}{(c) Denoise (Urban100+BSD68)} \\
\midrule
\tablehead
\multirow{2}{*}{Method} & \multirow{2}{*}{PSNR/SSIM} & \multirow{2}{*}{Method} & \multirow{2}{*}{PSNR/SSIM} & \multirow{2}{*}{Method} & \multicolumn{3}{c}{Urban100} & \multicolumn{3}{c}{BSD68} & \multirow{2}{*}{Average} \\
& & & & & $\sigma=15$ & $\sigma=25$ & $\sigma=50$ & $\sigma=15$ & $\sigma=25$ & $\sigma=50$ & \\
\midrule
DehazeNet~\citep{cai2016dehazenet} & 22.46/0.851 & DIDMDN~\citep{didmdn} & 23.79/0.773 & CBM3D~\citep{cbm3d} & 33.93/0.941 & 31.36/0.909 & 27.93/0.840 & 33.50/0.922 & 30.69/0.868 & 27.36/0.763 & 30.80/0.874 \\
MSCNN~\citep{mscnn} & 22.06/0.908 & UMR~\citep{umr} & 32.39/0.921 & DnCNN~\citep{DnCNN} & 32.98/0.931 & 30.81/0.902 & 27.59/0.833 & 33.89/0.930 & 31.23/0.883 & 27.92/0.789 & 30.74/0.878 \\
EPDN~\citep{epdn} & 22.57/0.863 & MSPFN~\citep{mspfn} & 33.50/0.948 & IRCNN~\citep{zhang2017learning} & 27.59/0.833 & 31.20/0.909 & 27.70/0.840 & 33.87/0.929 & 31.18/0.882 & 27.88/0.790 & 29.90/0.864 \\
FDGAN~\citep{fdgan} & 23.15/0.921 & LPNet~\citep{lpnet} & 33.61/0.958 & FFDNet~\citep{ffdnet} & 33.83/0.942 & 31.40/0.912 & 28.05/0.848 & 33.87/0.929 & 31.21/0.882 & 27.96/0.789 & 31.05/0.884 \\
AirNet~\citep{airnet} & 23.18/0.900 & AirNet~\citep{airnet} & 34.90/0.977 & BRDNet~\citep{brdnet} & 34.42/0.946 & 31.99/0.919 & 28.56/0.858 & 34.10/0.929 & 31.43/0.885 & 28.16/0.794 & 31.44/0.889 \\
Restormer~\citep{restormer} & 30.87/0.969 & Restormer~\citep{restormer} & 36.74/0.978 & AirNet~\citep{airnet} & 34.40/0.949 & 32.10/0.924 & 28.88/0.871 & 34.14/0.936 & 31.48/0.893 & 28.23/0.806 & 31.54/0.897 \\
PromptIR~\citep{promptir} & 31.31/0.973 & PromptIR~\citep{promptir} & 37.04/0.979 & PromptIR~\citep{promptir} & 34.77/0.952 & 32.49/0.929 & 29.39/0.881 & 34.34/0.938 & 31.71/0.897 & 28.49/0.813 & 31.87/0.902 \\
\alternativerow AdaIR~\citep{cui2025adair} & 31.80/0.981 & AdaIR~\citep{cui2025adair} & 38.90/0.985 & AdaIR~\citep{cui2025adair} & 34.96/0.953 & 32.74/0.931 & 29.70/0.885 & 34.36/0.938 & 31.72/0.897 & 28.49/0.813 & 32.00/0.903 \\
\methodrow \textbf{IRPO (Ours)} & \textbf{32.31/0.983} & \textbf{IRPO (Ours)} & \textbf{39.38/0.988} & \textbf{IRPO (Ours)} & \textbf{35.08/0.954} & \textbf{32.95/0.933} & \textbf{30.07/0.888} & \textbf{34.45/0.940} & \textbf{31.91/0.899} & \textbf{28.71/0.815} & \textbf{32.19/0.905} \\
\bottomrule
\end{tabular}
\end{adjustbox}
\end{table*}

\subsection{Additional Ablations}
Table~\ref{tab:grpo_groups} studies the rollout group size. Increasing $G$ from 3 to 4 improves the average result, while $G=5$ gives no clear gain and costs more rollouts. We therefore use $G=4$ in the main experiments. Table~\ref{tab:reward_source_comp} shows that Qwen-VL, Q-Align, and DeQA-Score lead to similar quantitative outcomes, suggesting that IRPO does not rely on a single expert model.

\begin{table}[t]
\centering
\caption{Ablation study on the GRPO group size ($G$). Performance is measured by average PSNR/SSIM on the all-in-one 5-task benchmark.}
\label{tab:grpo_groups}
\appendixtablefont
\begin{adjustbox}{max width=\columnwidth}
\begin{tabular}{@{}ccccccc@{}}
\toprule
\tablehead
Group & Dehazing & Deraining & Denoising & Deblurring & Low-Light & Average \\
\midrule
3 & 31.50/0.982 & 38.59/0.985 & 31.89/0.892 & 28.32/0.854 & 24.03/0.856 & 30.87/0.914 \\
\methodrow 4 & \textbf{31.85/0.984} & \textbf{38.72/0.987} & 32.13/0.896 & 28.89/0.863 & \textbf{24.08/0.858} & \textbf{31.13/0.918} \\
5 & 31.58/0.983 & 38.71/0.986 & \textbf{32.15/0.897} & \textbf{28.91/0.864} & 24.05/0.857 & 31.08/0.917 \\
\bottomrule
\end{tabular}
\end{adjustbox}
\end{table}

\begin{table}[t]
\centering
\caption{Comparisons of different expert reward sources on the all-in-one 5-task benchmark. Performance is measured by average PSNR/SSIM.}
\label{tab:reward_source_comp}
\appendixtablefont
\begin{adjustbox}{max width=\columnwidth}
\begin{tabular}{@{}lcccccc@{}}
\toprule
\tablehead
Expert Reward Model & Dehazing & Deraining & Denoising & Deblurring & Low-Light & Average \\
\midrule
w/ Q-Align~\citep{wu2024qalign} & 31.59/0.984 & 38.70/0.987 & 32.12/0.895 & 28.88/0.863 & 24.06/0.857 & 31.07/0.917 \\
w/ DeQA-Score~\citep{deqa_score} & 31.47/0.982 & 38.65/0.985 & 31.98/0.893 & 28.76/0.861 & 23.95/0.854 & 30.96/0.915 \\
\methodrow w/ Qwen2.5-VL~\citep{qwen25vl} & \textbf{31.85/0.984} & \textbf{38.72/0.987} & \textbf{32.13/0.896} & \textbf{28.89/0.863} & \textbf{24.08/0.858} & \textbf{31.13/0.918} \\
\bottomrule
\end{tabular}
\end{adjustbox}
\end{table}

\begin{table}[t]
\centering
\caption{Practical overhead of IRPO. The VLM reward is used only during post-training and is removed at inference.}
\label{tab:efficiency_main}
\appendixtablefont
\begin{adjustbox}{max width=\columnwidth}
\begin{tabular}{@{}lcccc@{}}
\toprule
\tablehead
Method & Params (M) & Post-Training Time & \makecell{Inference Time\\(1080p)} & \makecell{OOD Avg.\\PSNR} \\
\midrule
\alternativerow AdaIR & 28.78 & $\sim$14 GPU days pre-training & 2.0 s & 20.30 \\
\methodrow IRPO & 28.83 & +4.2 GPU days post-training & 2.0 s & \textbf{23.73} \\
\midrule
\textit{Difference} & \textit{+0.05 M (+0.17\%)} & \textit{one-time cost} & \textit{+0 ms} & \textit{\textbf{+3.43}} \\
\bottomrule
\end{tabular}
\end{adjustbox}
\end{table}

\subsection{Sensitivity Analysis of Reward Weights}
We analyze the reward weights $(\lambda_{gen}, \lambda_{expert}, \lambda_{task})$ on the Dehaze (SOTS) task. Table~\ref{tab:lambda_ablation_app} shows that the generic physical constraint $\lambda_{gen}$ should remain dominant ($\ge 0.5$) to preserve structural fidelity. The expert-quality weight $\lambda_{expert}$ works as a coarse visual regularizer and is kept relatively low ($\approx 0.1$) to avoid deviations from the ground truth. The setting $\lambda_{gen}=0.6$, $\lambda_{expert}=0.1$, and $\lambda_{task}=0.3$ gives the best balance in this study.

\begin{table}[t!]
    \centering
    \caption{Sensitivity analysis of the reward weights on the SOTS (Dehaze) dataset.}
    \label{tab:lambda_ablation_app}
    \appendixtablefont
    \begin{adjustbox}{max width=\columnwidth}
    \begin{tabular}{@{}ccccc@{}}
    \toprule
    \tablehead
    $\lambda_{gen}$ & $\lambda_{expert}$ & $\lambda_{task}$ & PSNR (dB) & SSIM \\
    \midrule
    0.3 & 0.4 & 0.3 & 31.45 & 0.980 \\
    0.5 & 0.2 & 0.3 & 32.12 & 0.981 \\
    \methodrow \textbf{0.6} & \textbf{0.1} & \textbf{0.3} & \textbf{32.31} & \textbf{0.983} \\
    0.7 & 0.1 & 0.2 & 32.24 & 0.983 \\
    \bottomrule
    \end{tabular}
    \end{adjustbox}
\end{table}

\subsection{Transfer to a Second Restoration Backbone}
\label{app:promptir_transfer}
We additionally evaluate the post-training recipe on the released PromptIR model~\citep{promptir}. We replace its three native routers with Dirichlet policy weights, keeping the prompt bases and encoder--decoder fixed. This adaptation uses a different policy parameterization from the Beta controls used for AdaIR. The post-training controls share the hard-data subset, 30 training epochs, crop settings, $\ell_1$/consistency supervision, and deterministic inference. The real-world evaluation covers RRSHID, GT-Rain, SIDD, ReLoBlur, and LoLi-Street.

Table~\ref{tab:promptir_transfer} reports the routing, GDM, and full IRPO variants. PromptIR-IRPO improves over the original PromptIR by 0.34~dB on the three-task average and 0.83~dB on the five real-world datasets. These results provide evidence of transfer beyond AdaIR with an architecture-specific policy interface.

\begin{table}[!htbp]
\centering
\caption{Post-training on PromptIR. Both columns report average PSNR/SSIM; the three-task setting follows the protocol in Table~\ref{tab:3D}.}
\label{tab:promptir_transfer}
\appendixtablefont
\begin{tabular}{@{}lcc@{}}
\toprule
Method & Three-task PSNR/SSIM & Five real-world PSNR/SSIM \\
\midrule
PromptIR & 32.06/0.913 & 21.71/0.687 \\
PromptIR + routing & 32.12/0.914 & 21.91/0.691 \\
PromptIR + GDM & 32.20/0.915 & 22.03/0.694 \\
\methodrow PromptIR-IRPO & \textbf{32.40/0.917} & \textbf{22.54/0.702} \\
\bottomrule
\end{tabular}
\end{table}

\section{Additional Visual Results}
\label{app:visual_results}

In addition to the synthetic comparison in Fig.~\ref{fig:synthetic_visual}, we provide real-world visual examples and more per-task qualitative comparisons against AirNet~\citep{airnet}, PromptIR~\citep{promptir}, and AdaIR~\citep{cui2025adair}.

\begin{figure*}[t!]
\centering
\includegraphics[width=0.9\textwidth]{real_world/real_world_horizontal.pdf}
\vspace{-0.2cm}
\caption{Visual comparisons on real-world datasets. From left to right: deblurring, dehazing, denoising, deraining, and low-light enhancement. IRPO reduces residual degradations while preserving local structures.}
\label{fig:real_world_app}
% \vspace{-8pt}
\end{figure*}

\begin{figure*}[t!]
\tabcolsep 0.8pt
\centering

\includegraphics[width=1\textwidth]{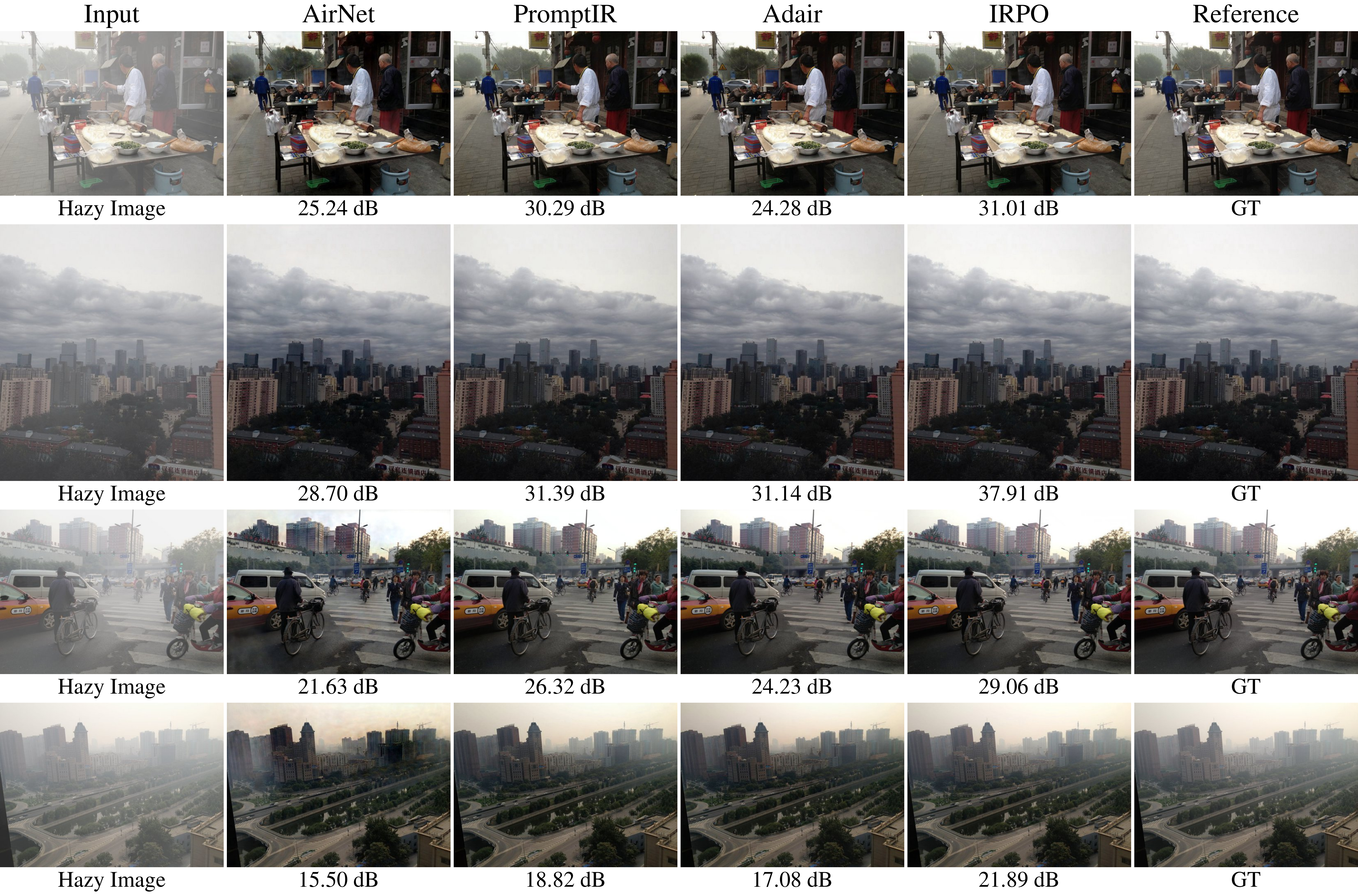}
% \vspace{-0.2cm}
\caption{Supplementary visual comparisons for image dehazing on synthetic datasets. IRPO yields clearer visibility and higher PSNR in these examples.}
\label{fig:supp_dehaze}
% \vspace{-12pt}
\end{figure*}

\noindent\textbf{Image Dehazing (Figure~\ref{fig:supp_dehaze}).}
Figure~\ref{fig:supp_dehaze} shows that competing methods can suffer from color distortion or residual haze. In the second row, AirNet and AdaIR leave dense haze in distant regions, leading to low contrast. IRPO recovers clearer visibility while maintaining more natural sky regions, which is consistent with the structural role of the \textit{General Reward}.

\begin{figure*}[t!]
\tabcolsep 0.8pt
\centering

\includegraphics[width=1\textwidth]{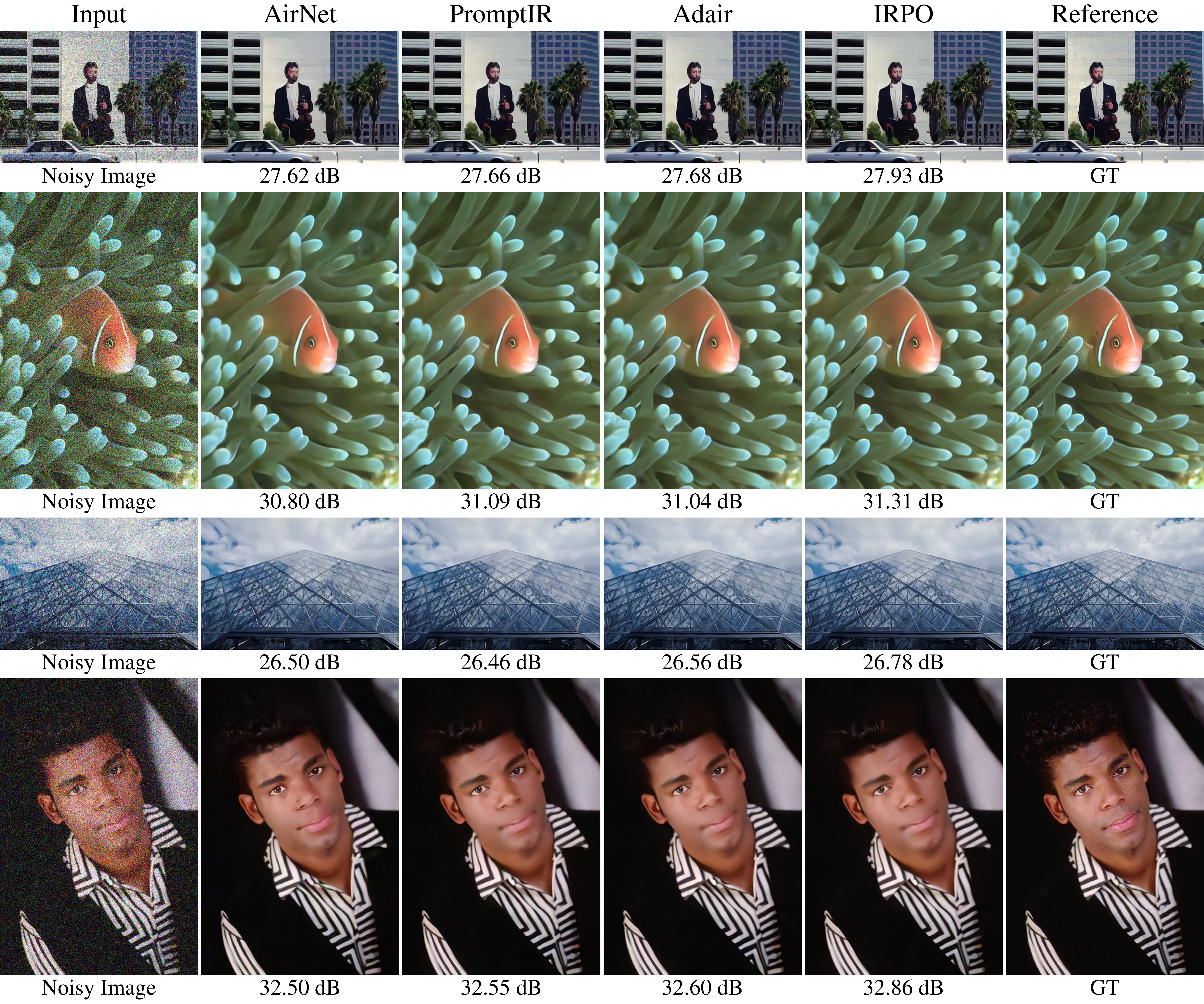}
% \vspace{-0.2cm}
\caption{Supplementary visual comparisons for image denoising (noise level 50) on synthetic datasets. IRPO preserves more fine details in these examples.}
\label{fig:supp_denoise}
% \vspace{-12pt}
\end{figure*}

\noindent\textbf{Image Denoising (Figure~\ref{fig:supp_denoise}).}
Figure~\ref{fig:supp_denoise} reports results at the $\sigma=50$ noise level. Severe noise often leads models to over-smooth details. In the "Face" example (fourth row), PromptIR and AdaIR blur facial features and hair textures. IRPO, guided by the \textit{Task-Aware Reward} based on gradient consistency, better balances noise removal and texture retention.

\begin{figure*}[t!]
\tabcolsep 0.8pt
\centering
\includegraphics[width=1\textwidth]{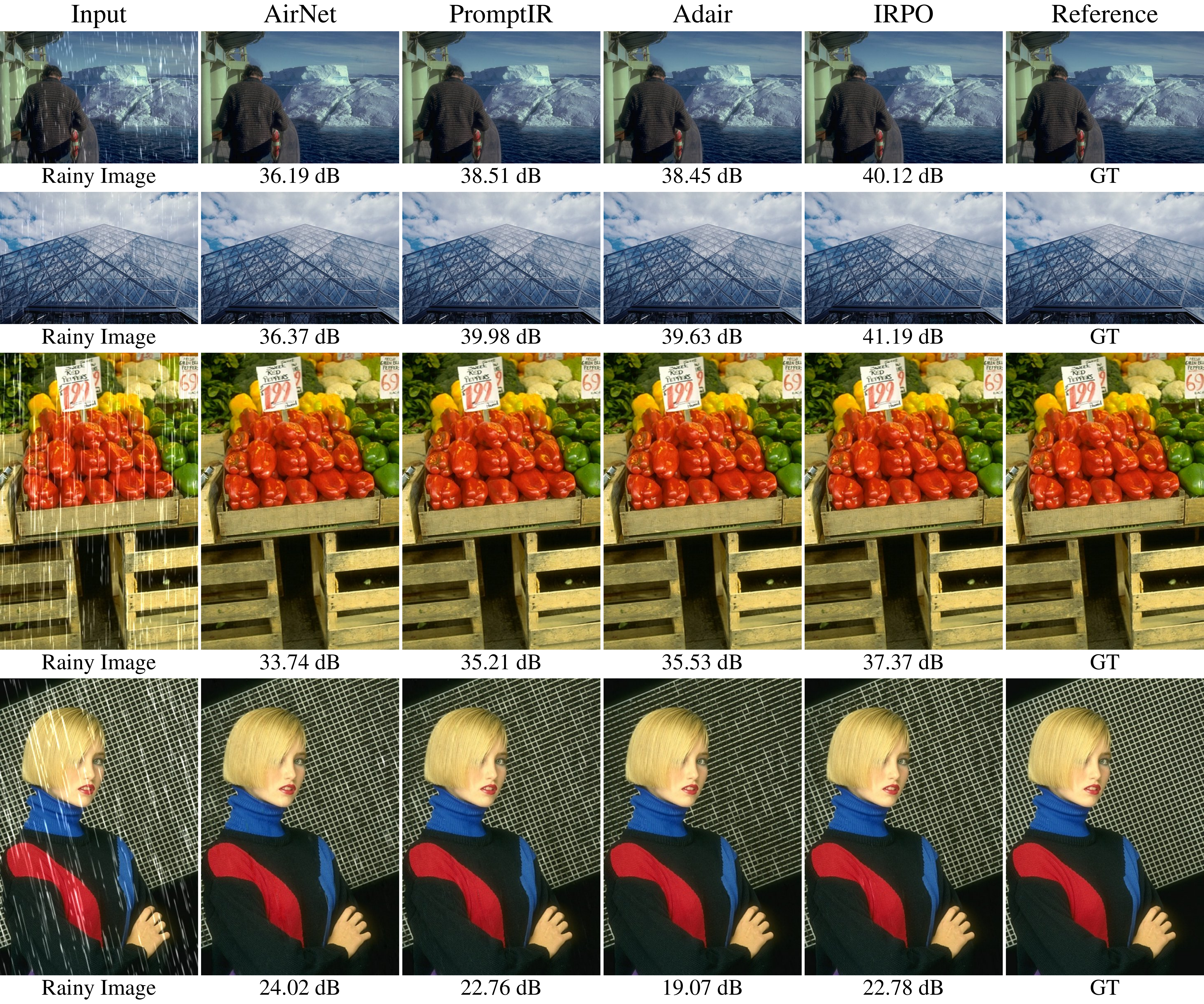}
% \vspace{-0.2cm}
\caption{Supplementary visual comparisons for image deraining on synthetic datasets. IRPO removes rain streaks while preserving background details.}
\label{fig:supp_derain}
% \vspace{-12pt}
\end{figure*}

\noindent\textbf{Image Deraining (Figure~\ref{fig:supp_derain}).}
Rain streaks often have clear directionality and varying density. In Figure~\ref{fig:supp_derain}, baselines leave rain artifacts in the first row and blur background details in the third row. IRPO uses the anisotropy consistency reward to separate rain streaks from background textures, leading to cleaner removal with fewer artifacts.

These qualitative comparisons are consistent with the quantitative results and support the data-oriented and reward-oriented design of IRPO.

\section{Additional Implementation Details}
\label{app:imp_details}

Table~\ref{tab:hyperparameters} lists the hyperparameters used in IRPO post-training. The table covers the model architecture, data handling, optimization, and GRPO training settings.

\begin{table}[t]
\centering
\footnotesize
\setlength{\tabcolsep}{4pt}
\renewcommand{\arraystretch}{1.15}
\caption{Detailed hyperparameters for IRPO post-training.}
\label{tab:hyperparameters}
\resizebox{0.8\columnwidth}{!}{
\begin{tabularx}{\columnwidth}{@{}L{.36\columnwidth} Y@{}}
\toprule
\textbf{Hyperparameter} & \textbf{Value / Description} \\
\midrule 
Base Channel Width (model\_dim) & 48 \\
Encoder-Decoder Levels & 4 levels \\
Transformer Blocks & [4, 6, 6, 8] \\
Multi-head Attention Heads & [1, 2, 4, 8] \\
Refinement Blocks & 4 \\
FFN Expansion Factor & 2.66 \\
Frequency Policy (AFLB) & 3 modules inserted between decoder levels \\
Action Space & 4D continuous $(r_h, r_l, g_f, g_o) \in (0,1)^4$ \\
Input Patch Size & $128 \times 128$ \\
Augmentation & Random horizontal/vertical flips \\
Hard-Mining Ratio & Top 30\% hardest samples \\
Optimizer & Adam ($\beta_1=0.9, \beta_2=0.999$) \\
Base Learning Rate (LR) & $3 \times 10^{-5}$ \\
Total Epochs & 30 (default for main results) \\
Batch Size & 1 per GPU (constrained by $G=4$) \\

Group Size ($G$) & 4 \\
Entropy Regularization & 0.01 \\
Supervised Loss Weight ($\lambda_{\text{sup}}$) & Linear decay: $0.35 \to 0.1$ \\
Consistency Loss Weight ($\lambda_{\text{cons}}$) & Linear decay: $0.2 \to 0.05$ \\
\bottomrule
\end{tabularx}
}
\end{table}

\section{Background on GRPO}
\label{app:grpo_details}

Our reward-oriented optimization pillar is based on Group Relative Policy Optimization (GRPO). Here, we provide the detailed formulation for completeness.

Given a context $x$ (the degraded image), a behavior policy $\pi_{\theta_{\text{old}}}(\cdot\mid x)$ draws a group of $G$ responses $\{a_i\}_{i=1}^{G}$ (actions, i.e., control parameters). Each corresponding output ${y}_i$ is scored by our reward model $R$ (Sec.~\ref{sec:reward_model}) to produce $\{r_i\}_{i=1}^{G}$.

GRPO constructs a \emph{group-normalized} relative advantage $A_i$ within the same input group, which serves as a stable, value-free advantage estimate:
\begin{align}
\bar r = \frac{1}{G}\sum_{i=1}^{G} r_i, \qquad
s = \sqrt{\frac{1}{G}\sum_{i=1}^{G}(r_i-\bar r)^2+\varepsilon}, \qquad
A_i = \frac{r_i-\bar r}{s}.
\label{eq:app_group_adv} % 
\end{align}
Let the likelihood ratio be:
\begin{equation}
\rho_i(\theta)=\frac{\pi_\theta(a_i\mid x)}{\pi_{\theta_{\text{old}}}(a_i\mid x)}.
\label{eq:app_ratio}
\end{equation}
GRPO then maximizes a clipped surrogate objective $\mathcal{J}_{\text{GRPO}}(\theta)$, which is identical to PPO but replaces the standard advantage with $A_i$. It further enforces a KL penalty on a reference policy $\pi_{\mathrm{ref}}$:
\begin{equation}
\label{eq:app_grpo_obj}
\begin{aligned}
\mathcal{J}_{\text{GRPO}}(\theta)= 
&\E_{x \sim \mathcal{D}, \{a_i\} \sim \pi_{\theta_{\text{old}}}}
\Bigg[
\frac{1}{G}\sum_{i=1}^{G}
\min\!\Big(
\rho_i(\theta)A_i,\;
\clip\!\big(\rho_i(\theta),1-\epsilon,1+\epsilon\big)A_i
\Big)
\\
&\qquad -\beta\,\KLhat\!\big(\pi_\theta(\cdot \mid x)\;\|\;\pi_{\mathrm{ref}}(\cdot \mid x)\big)
\Bigg].
\end{aligned}
\end{equation}
The KL divergence term is estimated on the batch using samples from $\pi_{\theta_{\text{old}}}$:
\begin{equation}
\begin{aligned}
\widehat{\mathcal{D}}_{\mathrm{KL}}\!\big(\pi_\theta\|\pi_{\mathrm{ref}}\big)
= 
\frac{1}{G} \sum_{i=1}^G \rho_i(\theta) \left[ \log \pi_\theta(a_i \mid x) - \log \pi_{\mathrm{ref}}(a_i \mid x) \right].
\label{eq:app_kl_est}
\end{aligned}
\end{equation}
The full objective used in the main paper, $\mathcal{L}_{\mathrm{RL}}$ (Eq.~\ref{eq:rl_loss}), is the negative of Eq.~\eqref{eq:app_grpo_obj} plus an entropy bonus. $A_i$ is treated as a stop-gradient statistic in practice.

\section{Reward Details}
\label{app:reward_details}

\noindent\textbf{Notation and utilities.}
We work with linear RGB in \([0,1]\). Let $\hat y$ denote the ground-truth image and $y$ the restored output. Let \(\mathrm{clip}_{[0,1]}(t)=\min(\max(t,0),1)\) and \(\varepsilon=10^{-6}\).
For luminance we use \(Y\!=\!0.299R+0.587G+0.114B\).
Sobel gradients of an image \(Z\) are \(G_x=\mathrm{Sobel}_x(Z)\), \(G_y=\mathrm{Sobel}_y(Z)\), and magnitude \(M(Z)=\sqrt{G_x^2+G_y^2+10^{-12}}\).
When needed, \(\mu(\cdot)\), \(\sigma^2(\cdot)\), and \(\mathrm{std}(\cdot)\) denote spatial mean, variance, and standard deviation.

% ==========================================================
% A.1: R_gen (THIS SECTION IS COMPLETELY REWRITTEN)
% ==========================================================
\subsection{Generic Quality Term \texorpdfstring{$R_{\text{gen}}$}{R\_gen}}
\label{app:generic_reward_details}
This reward is a weighted combination of five metrics, each normalized to \([0,1]\).
\begin{equation}
\begin{aligned}
R_{\text{gen}}(y, \hat y) = & \quad 0.25 \cdot R_{\text{clip}}(y, \hat y) + 0.25 \cdot R_{\text{LPIPS}}(y, \hat y) \nonumber + 0.15 \cdot R_{\text{aes}}(y)\\
&  + 0.20 \cdot R_{\text{PSNR}}(y, \hat y) \nonumber + 0.15 \cdot R_{\text{SSIM}}(y, \hat y),
\end{aligned}
\end{equation}
where $R_{\text{clip}}$ is the CLIP similarity~\citep{radford2021learning}, $R_{\text{LPIPS}}$ is the normalized LPIPS perceptual score~\citep{zhang2018unreasonable}, $R_{\text{aes}}$ is a normalized aesthetic score from a pre-trained predictor~\citep{schuhmann2022laion}, $R_{\text{PSNR}}$ is the PSNR score mapped to \([0,1]\) via fixed thresholds ($\tau_{\min}, \tau_{\max}$), and $R_{\text{SSIM}}$ is the standard SSIM score.

% ==========================================================
% A.2: R_qwen (PROMPT ADDED)
% ==========================================================
\subsection{VLM-based Verifier \texorpdfstring{$R_{\text{expert}}$}{R\_expert}}
\label{app:qwen_reward_details}
We use Qwen2.5-VL-7B-Instruct~\citep{qwen25vl} as a visual-quality verifier. The model receives the degraded input ($x$), the restored output ($y$), and the ground truth ($\hat y$). It returns a score from 1 to 5, which we linearly rescale.
The exact verifier prompt is provided at the end of this appendix section, after the reward definitions.

% ==========================================================
% A.3: R_task (JUSTIFICATIONS ADDED, FORMULAS VERIFIED)
% ==========================================================
\subsection{Task-aware Shaping \texorpdfstring{$R_{\text{task}}$}{R\_task}}
\label{app:task_reward_details}
For each degradation type, we define a specialized reward.

\paragraph{(1) Denoising --- gradient consistency.}
Let \(M_{g}=M(Y_{g})\), \(M_{\text{rest}}=M(y)\).
Define the GT baseline and mean magnitude deviation:
\begin{equation}
B=\mu\!\big(M_{g}\big),\qquad
D=\mu\!\big(|M_{\text{rest}}-M_{g}|\big).
\end{equation}
The reward encourages the restored gradients to approach the clean target:
\begin{equation}
\label{eq:r_grad}
R_{\text{grad}}=\mathrm{clip}_{[0,1]}\!\left(1-\frac{D}{B+\varepsilon}\right).
\end{equation}
This reward encourages the restored texture and edge gradients ($M_{\text{rest}}$) to match the ground truth ($M_{g}$), reducing over-smoothing during denoising.

\paragraph{(2) Deraining --- anisotropy consistency.}
Compute average absolute Sobel responses on luminance:
\begin{equation}
E_x=\mu\!\big(|G_x(y)|\big),\qquad
E_y=\mu\!\big(|G_y(y)|\big).
\end{equation}
Define normalized anisotropy
\begin{equation}
A=\frac{|E_x-E_y|}{\max(E_x+E_y,\varepsilon)}.
\end{equation}
The reward favors isotropy (i.e., $A \approx 0$):
\begin{equation}
\label{eq:r_aniso}
R_{\text{aniso}}=1-\mathrm{clip}_{[0,1]}(A).
\end{equation}
Rain streaks are anisotropic and create a directional bias in image gradients ($E_x \neq E_y$). A clean, rain-free image is closer to isotropic ($E_x \approx E_y$). This reward encourages isotropic outputs after deraining.

\paragraph{(3) Dehazing --- contrast closeness.}
Let \(C(Z)=\mathrm{std}(Y_Z)\) be a proxy for the contrast of image \(Z\).
We measure symmetry-aware contrast closeness to ground truth:
\begin{equation}
\label{eq:r_contrast}
R_{\text{contrast}}=\mathrm{clip}_{[0,1]}\!\left(\min\!\left(\frac{C_{\text{rest}}}{C_{g}+\varepsilon},\;
\frac{C_{g}}{C_{\text{rest}}+\varepsilon}\right)\right).
\end{equation}
Haze is a low-frequency component that reduces image contrast (lower $\mathrm{std}(Y)$). This reward encourages the restored contrast $C_{\text{rest}}$ to match the ground truth $C_{g}$, penalizing both under-enhancement (remaining haze) and over-enhancement (oversaturation).

\paragraph{(4) Deblurring --- sharpness closeness.}
Using the mean gradient magnitude \(S(Z)=\mu(M(Z))\) as a proxy for sharpness:
\begin{equation}
\label{eq:r_sharp}
R_{\text{sharp}}=\mathrm{clip}_{[0,1]}\!\left(\min\!\left(\frac{S_{\text{rest}}}{S_{g}+\varepsilon},\;
\frac{S_{g}}{S_{\text{rest}}+\varepsilon}\right)\right).
\end{equation}
Blur removes high-frequency information and lowers gradient magnitudes. This reward encourages the mean sharpness $S_{\text{rest}}$ to match the ground truth $S_{g}$, which supports edge and detail recovery.

\paragraph{(5) Low-light --- exposure \& color consistency.}
Exposure alignment on luminance means:
\begin{equation}
d=\big|\mu(y)-\mu(Y_{g})\big|,\qquad
R_{\text{exp}}=\mathrm{clip}_{[0,1]}\!\left(1-\frac{d}{0.5}\right).
\end{equation}
Color consistency sums mean-channel deviations (\(c\in\{R,G,B\}\)):
\begin{equation}
d_c=\sum_{c}\big|\mu(c_{\text{rest}})-\mu(c_{g})\big|,\qquad
R_{\text{color}}=\mathrm{clip}_{[0,1]}\!\left(1-\frac{d_c}{0.6}\right).
\end{equation}
Low-light enhancement has two goals: fidelity and brightness. $R_{\text{exp}}$ ensures the overall luminance $\mu(y)$ matches the target, while $R_{\text{color}}$ prevents color shifts (e.g., a green or purple tint) by penalizing deviation in individual R, G, B channels.

\paragraph{Putting the task terms together.}
For each sample, we select the reward(s) corresponding to its degradation:
\begin{equation}
\label{eq:rtask_final}
\begin{gathered}
R_{\text{task}}^{(\text{denoise})} = R_{\text{grad}}, \quad
R_{\text{task}}^{(\text{derain})} = R_{\text{aniso}}, \quad\\
R_{\text{task}}^{(\text{dehaze})} = R_{\text{contrast}},
R_{\text{task}}^{(\text{deblur})} = R_{\text{sharp}}, \\
R_{\text{task}}^{(\text{low-light})}
 = 0.2 \cdot R_{\text{exp}}
 + 0.1 \cdot R_{\text{color}}.
\end{gathered}
\end{equation}

% ==========================================================
% A.4: Final Combination (NEW SECTION)
% ==========================================================
\subsection{Final Reward Combination}
The final reward signal used in Eq.~(\ref{eq:reward_main}) of the main paper is the weighted sum of the three components:
\begin{align}
R(y, \hat y) =  \quad 0.6 \cdot R_{\text{gen}}(y, \hat y) \nonumber  + 0.1 \cdot R_{\text{expert}}(y) \nonumber  + 0.3 \cdot R_{\text{task}}^{(k)}(y, \hat y)
\end{align}
where $k$ is the known degradation type for the training sample.
% with fixed \(\lambda_1,\lambda_2,\lambda_3\).

\subsection{VLM Verifier Prompt}
The following prompt is used to obtain the expert score before linear rescaling.
\begin{tcolorbox}[
    width=\columnwidth,
    colback=gray!5,
    colframe=gray!75!black,
    fonttitle=\bfseries,
    title=Qwen-VL Verifier Prompt
]
\small

You are an image-restoration expert. You will be given three images:
\begin{enumerate}
    \item The degraded input that suffers from a certain type of degradation.
    \item The restored output generated by a model.
    \item The clean ground-truth reference.
\end{enumerate}

\textbf{Task:}
\begin{enumerate}
    \item Identify the most plausible degradation type of the input image. Consider categories such as denoising (0/1/2, different noise levels), deraining (3), dehazing (4), deblurring (5), or low-light enhancement (6). Briefly justify your reasoning.
    \item Compare the restored output against the ground truth with respect to the identified degradation type. Pay attention to:
    \begin{itemize}
        \item Noise or streak removal quality for denoising/deraining.
        \item Contrast and haze removal for dehazing.
        \item Sharpness recovery for deblurring.
        \item Exposure and color constancy for low-light enhancement.
    \end{itemize}
    \item Highlight specific improvements and any remaining artifacts.
    \item Provide a final quality score from 1 to 5, where:
    \begin{itemize}
        \item 1: severe artifacts or almost no improvement,
        \item 2: minor improvement but significant issues remain,
        \item 3: moderate improvement with noticeable gaps,
        \item 4: good restoration with only small flaws,
        \item 5: near-perfect restoration indistinguishable from ground truth.
    \end{itemize}
\end{enumerate}
Respond in the following XML-style format:
\begin{verbatim}
<Assessment>
  <Degradation>
    [type and reasoning]
  </Degradation>
  <Analysis>
    [detailed comparison]
  </Analysis>
  <Score>X</Score>
</Assessment>
\end{verbatim}
\end{tcolorbox}

\section{Policy Parametrization Details}
\label{app:policy_details}

Our policy $\pi_\theta$ consists of 6 independent policy heads distributed across the 3 AFLB modules in the AdaIR decoder. Each AFLB module contains two heads: (1) a \textbf{Policy Rate} head, and (2) a \textbf{Policy Fuse} head, which correspond to the actions $(r_h, r_l)$ and $(g_f, g_o)$ respectively.

\paragraph{Policy Rate Head $(r_h, r_l)$.}
The Policy Rate head determines the size of the frequency-domain mask.
\begin{itemize}
    \item \textbf{Input:} It takes the initial degraded image $x$ (which is fed to all AFLB modules) and passes it through an FFT.
    \item \textbf{Process:} A small CNN head takes the pooled features of $x$ and outputs four positive scalars $(\alpha_h, \beta_h, \alpha_l, \beta_l)$.
    \item \textbf{Action:} These scalars parameterize two Beta distributions. During training, we sample $r_h \sim \mathrm{Beta}(\alpha_h,\beta_h)$ and $r_l \sim \mathrm{Beta}(\alpha_l,\beta_l)$. These actions $(r_h, r_l) \in (0,1)^2$ define the height and width of the low-frequency mask. This mask is applied to the FFT of $x$ to separate frequency components, which are then used to generate the intermediate feature $O_i$ (the "Output Feature" in Figure~\ref{fig:pipeline}).
\end{itemize}

\paragraph{Policy Fuse Head $(g_f, g_o)$.}
The "Policy Fuse" head determines how to blend the intermediate feature with the decoder's latent feature.
\begin{itemize}
    \item \textbf{Input:} It takes the latent feature $y_i$ (the "Latent Feature" in Figure~\ref{fig:pipeline}) coming from the previous decoder block.
    \item \textbf{Process:} A separate small CNN head takes the pooled features of $y_i$ and outputs four positive scalars $(\alpha_1, \beta_1, \alpha_2, \beta_2)$.
    \item \textbf{Action:} We sample $g_f \sim \mathrm{Beta}(\alpha_1,\beta_1)$ and $g_o \sim \mathrm{Beta}(\alpha_2,\beta_2)$. These actions $(g_f, g_o) \in (0,1)^2$ are the dynamic weights used in the final fusion, as shown in Eq.~\eqref{eq:aflb_fusion} in the main paper.
\end{itemize}

\paragraph{Log-Probability Calculation.}
The policy $\pi_\theta$ is the joint distribution of these 4 independent actions. The log-probability $\log \pi_\theta(a\!\mid\! x, y)$ is required for the policy gradient calculation (e.g., in Eq.~\ref{eq:app_kl_est}). This function has the following closed form:
\begin{equation}
\label{eq:beta_logprob_tight}
\begin{aligned}
\log \pi_\theta(a\!\mid\! x, y)
&= \sum_{k \in \{h,l,f,o\}} \Bigl[(\alpha_k\!-\!1)\,\log a_k + (\beta_k\!-\!1)\,\log\bigl(1-a_k\bigr)\Bigr] \\
&\quad -\, \sum_{k \in \{h,l,f,o\}} \log B(\alpha_k,\beta_k),
\end{aligned}
\end{equation}
where $a_k$ is the sampled action for $k \in \{h, l, f, o\}$, $(\alpha_k, \beta_k)$ are the distribution parameters output by the corresponding policy head, and $B(\cdot,\cdot)$ is the Beta function.

\clearpage
\section*{NeurIPS Paper Checklist}

\begin{enumerate}

\item {\bf Claims}
    \item[] Question: Do the main claims made in the abstract and introduction accurately reflect the paper's contributions and scope?
    \item[] Answer: \answerYes{}
    \item[] Justification: The abstract and introduction state IRPO as a GRPO-based post-training framework for image restoration, with data-oriented supervision, reward-oriented optimization, and empirical gains on in-domain and OOD benchmarks. The claims are framed as fidelity-oriented and quality-aware restoration rather than direct human-preference alignment.

\item {\bf Limitations}
    \item[] Question: Does the paper discuss the limitations of the work performed by the authors?
    \item[] Answer: \answerYes{}
    \item[] Justification: The main text discusses the scope of the post-training wrapper, the AdaIR-specific stochastic-head parameterization, the one-time VLM-assisted post-training cost, and the role of IQA metrics as supporting diagnostics rather than a replacement for human preference studies.

\item {\bf Theory assumptions and proofs}
    \item[] Question: For each theoretical result, does the paper provide the full set of assumptions and a complete (and correct) proof?
    \item[] Answer: \answerNA{}
    \item[] Justification: The paper is an empirical method paper and does not claim new theoretical guarantees. The GRPO objective and its instantiation for restoration are described for completeness.

\item {\bf Experimental result reproducibility}
    \item[] Question: Does the paper fully disclose all the information needed to reproduce the main experimental results of the paper to the extent that it affects the main claims and/or conclusions of the paper (regardless of whether the code and data are provided or not)?
    \item[] Answer: \answerYes{}
    \item[] Justification: The paper reports datasets, evaluation protocols, metrics, baselines, model initialization, optimization settings, group size, reward components, and hyperparameters. Additional implementation details and ablations are provided in the appendix.

\item {\bf Open access to data and code}
    \item[] Question: Does the paper provide open access to the data and code, with sufficient instructions to faithfully reproduce the main experimental results, as described in supplemental material?
    \item[] Answer: \answerNo{}
    \item[] Justification: The experiments use public benchmark datasets and cite their original sources. A public code URL is not provided in this manuscript.

\item {\bf Experimental setting/details}
    \item[] Question: Does the paper specify all the training and test details (e.g., data splits, hyperparameters, how they were chosen, type of optimizer) necessary to understand the results?
    \item[] Answer: \answerYes{}
    \item[] Justification: The experimental section specifies the in-domain and OOD datasets, all-in-one and single-task settings, evaluation metrics, optimizer, learning rate, crop size, group size, training epochs, reward weights, and compute setup. The appendix lists detailed hyperparameters.

\item {\bf Experiment statistical significance}
    \item[] Question: Does the paper report error bars suitably and correctly defined or other appropriate information about the statistical significance of the experiments?
    \item[] Answer: \answerNo{}
    \item[] Justification: Following common image-restoration benchmark practice, the main tables report deterministic benchmark PSNR/SSIM and IQA scores rather than repeated-run confidence intervals. Training stability curves are included, but repeated-seed error bars for every benchmark are not reported.

\item {\bf Experiments compute resources}
    \item[] Question: For each experiment, does the paper provide sufficient information on the computer resources (type of compute workers, memory, time of execution) needed to reproduce the experiments?
    \item[] Answer: \answerYes{}
    \item[] Justification: The implementation details specify the GPU setup used for IRPO post-training and VLM reward serving, and the main paper reports post-training time, parameter overhead, and inference latency.

\item {\bf Code of ethics}
    \item[] Question: Does the research conducted in the paper conform, in every respect, with the NeurIPS Code of Ethics?
    \item[] Answer: \answerYes{}
    \item[] Justification: The work studies image-restoration training on public benchmarks, does not collect private user data, and does not conduct human-subject experiments. The paper avoids unsupported claims of human-preference alignment.

\item {\bf Broader impacts}
    \item[] Question: Does the paper discuss both potential positive societal impacts and negative societal impacts of the work performed?
    \item[] Answer: \answerYes{}
    \item[] Justification: The method can improve restoration robustness for images captured under adverse degradations, which may benefit downstream visual analysis. Potential negative uses include over-reliance on enhanced imagery or deployment in sensitive surveillance settings; the method should be used with task-specific validation and provenance awareness.

\item {\bf Safeguards}
    \item[] Question: Does the paper describe safeguards that have been put in place for responsible release of data or models that have a high risk for misuse?
    \item[] Answer: \answerNA{}
    \item[] Justification: The paper does not introduce a high-risk dataset or a model designed for harmful generation. It uses public restoration benchmarks and releases a restoration training recipe rather than private data or identity-recognition capability.

\item {\bf Licenses for existing assets}
    \item[] Question: Are the creators or original owners of assets (e.g., code, data, models), used in the paper, properly credited and are the license and terms of use explicitly mentioned and properly respected?
    \item[] Answer: \answerYes{}
    \item[] Justification: The paper cites the original datasets, baselines, metrics, and Qwen2.5-VL source used for expert reward scoring. Public assets are used for research benchmarking under their respective release terms.

\item {\bf New assets}
    \item[] Question: Are new assets introduced in the paper well documented and is the documentation provided alongside the assets?
    \item[] Answer: \answerNo{}
    \item[] Justification: The paper documents the method, hyperparameters, reward components, and evaluation protocols, but a public code and configuration release is not linked in this manuscript.

\item {\bf Crowdsourcing and research with human subjects}
    \item[] Question: For crowdsourcing experiments and research with human subjects, does the paper include the full text of instructions given to participants and screenshots, if applicable, as well as details about compensation (if any)?
    \item[] Answer: \answerNA{}
    \item[] Justification: The paper does not include crowdsourcing studies or direct human-subject experiments.

\item {\bf Institutional review board (IRB) approvals or equivalent for research with human subjects}
    \item[] Question: Does the paper describe potential risks incurred by study participants, whether such risks were disclosed to the subjects, and whether IRB approvals were obtained?
    \item[] Answer: \answerNA{}
    \item[] Justification: No human-subject protocol is conducted in this study.

\item {\bf Declaration of LLM usage}
    \item[] Question: Does the paper describe the usage of LLMs if it is an important, original, or non-standard component of the core methods in this research?
    \item[] Answer: \answerYes{}
    \item[] Justification: The paper explicitly describes Qwen2.5-VL as a train-time expert reward model. It is used only to score restoration candidates during post-training and is removed at inference.

\end{enumerate}

\end{document}